\documentclass[runningheads]{llncs}
\usepackage{graphicx}
\usepackage{comment}
\usepackage{amsmath,amssymb} 
\usepackage{color}
\usepackage{multirow}
\usepackage[table]{xcolor}
\usepackage{hyperref}
\hypersetup{
	colorlinks=true,
	linkcolor=cyan,
	filecolor=blue,      
	urlcolor=red,
	citecolor=green,
}
\newcommand{\edit}{\textcolor{black}}
\newcommand{\etal}[0]{\textit{et al.}}
\newcommand{\bg}{\boldsymbol{\alpha}}
\newcommand{\bs}{\boldsymbol{\beta}}
\newcommand{\bp}{\boldsymbol{\theta}}

\begin{document}
\pagestyle{headings}
\mainmatter
\def\ECCVSubNumber{3361}  

\title{BCNet: Learning Body and Cloth Shape from A Single Image} 

\titlerunning{BCNet}
%
\author{Boyi Jiang\inst{1}, Juyong Zhang\inst{1}\thanks{Corresponding author. Email: {\texttt{juyong@ustc.edu.cn}}.}, Yang Hong\inst{1},  Jinhao Luo\inst{1}, Ligang Liu\inst{1}, \and Hujun Bao\inst{2}}

\authorrunning{B. Jiang et al.}
%
\institute{University of Science and Technology of China \and
State Key Lab of CAD\&CG, Zhejiang University}

\maketitle

\begin{abstract}
In this paper, we consider the problem to automatically reconstruct garment and body shapes from a single near-front view RGB image. To this end, we propose a layered garment representation on top of SMPL and novelly make the skinning weight of garment independent of the body mesh, which significantly improves the expression ability of our garment model. Compared with existing methods, our method can support more garment categories and recover more accurate geometry. To train our model, we construct two large scale datasets with ground truth body and garment geometries as well as paired color images. Compared with single mesh or non-parametric representation, our method can achieve more flexible control with separate meshes, makes applications like re-pose, garment transfer, and garment texture mapping possible. Code and some data is available at \href{https://github.com/jby1993/BCNet}{https://github.com/jby1993/BCNet}.

\end{abstract}

\section{Introduction}
Applications like virtual try-on, VR/AR, and entertainment need detailed and accurate reconstruction of both body and dressed garments with simple input like color image. However, the variety of body shapes, postures and garment categories, makes it a very challenging problem. A simulation-based method~\cite{yang2018physics} explores this problem, but their solution is dedicated and time-consuming. In this paper, we aim to automatically reconstruct both body and cloth shapes from just a single near-front view image, utilizing the powerful fitting ability of the deep neural network.

In recent years, body shape reconstruction from images has made significant progress~\cite{kanazawa2018end,omran2018neural,kanazawa2019learning,liu2019temporally}.
A common way is to infer the shape and pose parameter of a statistical body model, like SMPL~\cite{loper2015smpl}. These methods are robust for different posture, but the reconstructed geometry is constrained to be within the model space, which can not capture the complex cloth shape.

To infer detailed geometry beyond body shape, some non-parametric representations have been proposed~\cite{varol2018bodynet,zheng2019deephuman,natsume2019siclope,saito2019pifu}. These non-parametric representations based on voxel and implicit function can recover arbitrary shapes. However, voxel representation is hard to recover shape details due to their large memory consumption for high resolution. Although implicit representation is more memory efficient, it may generate infeasible results like broken arms. Moreover, the lack of semantic information limits their applications like garment transfer.

Expanding the representation ability of the statistical model of body shape is another solution. Several prior works~\cite{bhatnagar2019multi,pavlakos2018learning,alldieck2018video,alldieck2019tex2shape} utilize the vertex displacements of body shape represented by SMPL to represent garment geometry. Under this configuration, tight garments can be reconstructed. However, this representation cannot recover the feature of garment edges. More importantly, binding garments with SMPL points causes the problem that garments have the same skinning weights and connectivity with SMPL. Therefore, large scale displacements of loose garments may cause artifacts because of inappropriate skinning weights. More importantly, garments like skirts which have a different topology with body shape, are beyond the representation range.

Like Bhatnagar \etal~\cite{bhatnagar2019multi}, we train a model to reconstruct body mesh and layered garment meshes separately. \edit{The difference in input is that our method only requires a single RGB image and no additional semantic information and body rough A-pose constrain.} Another difference is that our garment mesh is not bound with the body mesh, and can reconstruct more garment categories. To this end, we address three major challenges: learning a shared skinning weight network for all garments, garment detail inference, and dataset construction. Our method supports six garment categories, including upper garment, pants, and skirts with short and long templates for each type. For all garment types, we train a network to predict skinning weights related to SMPL's skeleton. For each type, we use graph convolution to recover the details. To train the model, a dataset with various RGB images and their corresponding body and cloth shapes is needed. However, there is no available public datasets that satisfy our demands. Instead, for each type of garment, we design different sizes of clothes dressed on different SMPL neutral bodies and repose these clothes to various postures utilizing a physics engine. Besides, a commercial 3D human dataset with high-definition texture is added to increase the diversity of training data.

Our method can infer both body and garment shapes from a single image with different poses, and also supports loose garment types, like skirts. Based on the reconstruction results by our method, applications like garments and poses transferring between different images can be achieved. In summary, the contributions of this work include the following aspects:
\begin{itemize}
\item We present a novel garment representation on top of SMPL and a neural network-based method to reconstruct the shapes of body and garment from a single near-front viewpoint color image.
\item Rather than binding the skinning weight of garment with body mesh, we propose a generic skinning weights generating network, which enables our approach to support garments with different topologies.
\item We design a complete algorithm pipeline for dressed SMPL body data construction with different types of garments. The constructed dataset, including synthetic images and clothed body shapes, will be publicly available.
\end{itemize}
\section{Related Work}

\noindent{\bf Template-Free Clothed Human Estimation.} Some non-parametric methods based on voxel or implicit function have been proposed to address the complex topology of garments. BodyNet~\cite{varol2018bodynet} directly infers a voxel representation of clothed bodies with a deep network. Due to the large memory cost for high resolution, high-frequency details are often missed. Jackson \etal~\cite{jackson20183d} reconstruct the shape of humans via volumetric regression and show the ability to output fine-scale details. Zheng \etal~\cite{zheng2019deephuman} infer clothed body volume representation with an initial aligned SMPL body, and combine image features to enhance reconstruction details. Natsume \etal~\cite{natsume2019siclope} propose a reconstruction method based on a multi-view framework using synthesizing new silhouettes from a single image. More recently, ~\cite{saito2019pifu} proposes a promising clothed body reconstruction network using a memory-efficient implicit representation. Template-free methods do not utilize the human body prior to obtain complex topology modeling ability, at the cost of lacking semantic information and control of reconstructed results.

\noindent{\bf Template-Based Clothed Human Estimation.} Based on human body statistical model~\cite{loper2015smpl,anguelov2005scape,jiang2019learning}, many works can estimate naked body shape from image~\cite{kanazawa2018end,lassner2017unite,pavlakos2018learning,omran2018neural,bogo2016keep,dibra2017human,yao2019densebody}.  For better representation ability, a displacement vector is added for each vertex.~\cite{alldieck2018video,alldieck2019learning,alldieck2018detailed} adopt this strategy to reconstruct clothed body with skin-tight garment. Alldieck \etal~\cite{alldieck2019tex2shape} estimate detailed normal and vector displacement on the UV map, which leads to finer-scale details. Zhu \etal~\cite{zhu2019detailed} model fine-scale details by adding free-form 3D deformation on top of parametric model. Instead of using a single surface to represent both garment and body, ~\cite{bhatnagar2019multi} separates SMPL mesh to represent upper garment and pant independently, leading to more flexible control. However, binding garment vertices to body model strictly restricts the topology of support garment categories, and it is hard to represent more loose garment types, such as skirts. \edit{~\cite{pons2017clothcap,yu2019simulcap} also use separate body and garment templates to register clothed body motion sequences.}

\noindent{\bf Garment Dataset Construction.} BUFF~\cite{zhang2017detailed} supplies high-quality 4D scans of clothed bodies, \edit{but it only has 5 subjects and 2 suits for each subject}. Lahner~\etal~\cite{lahner2018deepwrinkles} collect high-quality 4D scans of garments, but the method leaves out body reconstruction, and their dataset is not publicly available. Recently, \cite{bhatnagar2019multi} constructs a training dataset with garment and body shapes from real scan data, but the training dataset is also unavailable. Moreover, many prior works generate ground truth dataset based on physics-based simulation~\cite{liu2019temporally,liang2019shape,santesteban2019learning,wang2018learning,danvevrek2017deepgarment,guan2012drape,pumarola20193dpeople}.
~\cite{liu2019temporally,liang2019shape} dress SMPL bodies and construct more truthful images than SURREAL~\cite{varol2017learning}. ~\cite{wang2018learning} simulates three types of garment and dress them on neutral SMPL bodies to learn garments design from sketches. All mentioned datasets do not meet our requirements. Therefore, we build a dataset containing a variety of garments and body types with different sizes and postures.  

\noindent{\bf Garment Deformation Representation.} How to represent the deformation of a garment is also related to our work. De Aguiar \etal~\cite{de2010stable} represent the garment dynamic dressed on a specific virtual avatar with a linear combination of pre-computed multiple deformations. DRAPE~\cite{guan2012drape} regresses garment deformation from body shape with a technique derived from SCAPE~\cite{anguelov2005scape}. Xu \etal~\cite{xu2014sensitivity} combine rotation and translation weights
to approximate the non-local and nonlinear clothing deformation and introduce a pose sensitive rigging scheme. Lahner \etal~\cite{lahner2018deepwrinkles} recover high-frequency garment details from a normal map created from Generative Adversarial Network. \edit{Yang \etal~\cite{yang2018analyzing} model garments with different connectivity based on a body template and use PCA to parameterize garment deformation. Santesteban \etal~\cite{santesteban2019learning} propose to deform base garment conditioned on body parameters and then add high-frequency wrinkles.} 
\begin{figure*}[t!]
\begin{center}
\includegraphics[width=\linewidth]{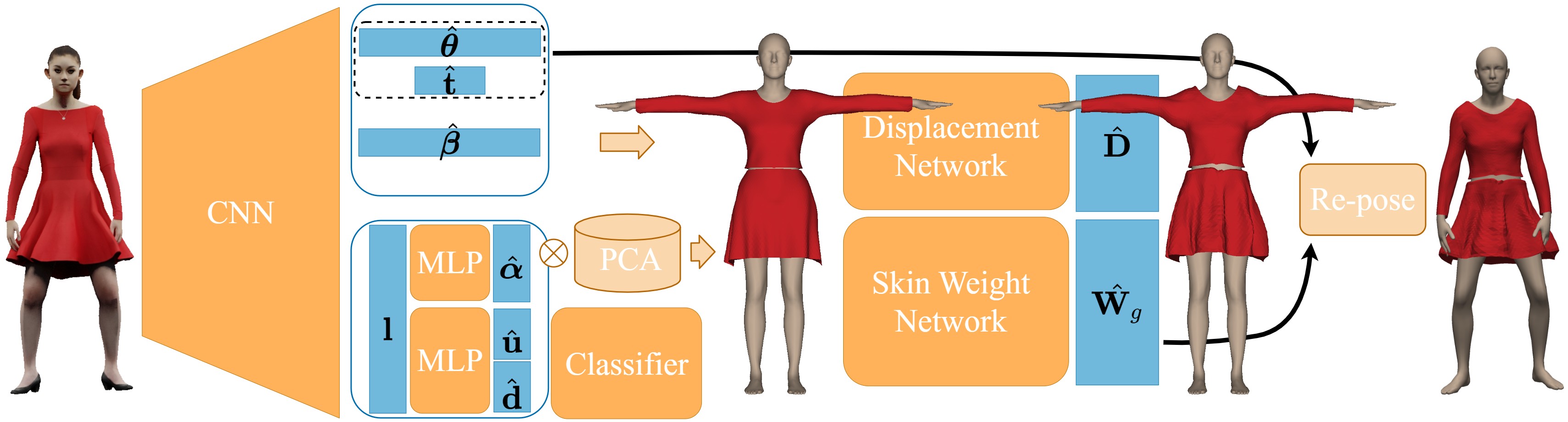}
\end{center}
\caption{The architecture of our proposed network. The CNN encodes image into latent feature, then we get reconstructed SMPL parameters $\hat{\bs}$, $\hat{\bp}$, $\hat{\mathbf{t}}$ and shared garment latent feature $\mathbf{l}$ with respective FC layers. From $\mathbf{l}$, we reconstruct garment shape parameter $\hat{\bg}$ and garment type scores $\{\hat{\mathbf{u}},\hat{\mathbf{d}}\}$ for upper and lower garment separately. With the classifier, $\hat{\bg}$ and $\hat{\bs}$, we reconstruct neutral clothed body. Followed a displacement network and skinning weight network, we predict garment vertex displacements and skinning weights separately. Finally, utilizing predicted pose parameters $\hat{\bp}$, $\hat{\mathbf{t}}$ and $\hat{\mathbf{W}_g}$, we re-pose neutral body and garments with displacements to reference posture.}
\label{fig:pipeline}
\end{figure*}

\section{Algorithm}
The target of this work is to automatically reconstruct both body and cloth shapes from a single near-front view image. Our model currently supports six garment categories and can be easily extended to other new types. In the following, we first describe our garment representation model. Then, we introduce our network structure and training loss design.

\subsection{Garment Model}
\label{sec:gar_model}
We use SMPL~\cite{loper2015smpl} as our parametric human body model. SMPL is a function which maps shape parameters $\bs\in{\mathbb{R}^{10}}$ and pose parameters $\bp\in{\mathbb{R}^{72}}$ to a body mesh $\mathbf{M}_b(\bs,\bp)\in{\mathbb{R}^{3|\mathcal{V}_b|}}$, where $\mathcal{V}_b$ is SMPL mesh vertices set. The mapping can be summarized as the following equation:
\begin{equation}
\mathbf{M}_b(\bs,\bp)=\mathit{W}(\mathbf{T}_b(\bs,\bp),\mathbf{J}(\bs),\bp,\mathbf{W}_b), \quad
\mathbf{T}_b(\bs,\bp)=\mathbf{B}+\mathbf{B}_s\bs+\mathbf{B}_p\bp,
\end{equation}
where SMPL applies linear displacement bases $\mathbf{B}_s$ and $\mathbf{B}_p$ on a T-posed template mesh $\mathbf{B}$, and then utilize standard skeleton skinning operation $\mathit{W}$ to get posed body mesh. $\mathbf{J}(\bs)\in{\mathbb{R}^{24\times3}}$ is SMPL body's neutral skeleton and $\mathbf{W}_b\in{\mathbb{R}^{|\mathcal{V}_b|\times24}}$ is the skinning weights of each vertex of SMPL.

As most clothes follow the deformations of the body, we compute our garment mesh $\mathbf{M}_g\in{\mathbb{R}^{3|\mathcal{V}_g|}}$ similarly based on the skin deformation of SMPL:
\begin{equation}
\mathbf{M}_g(\bg,\bs,\bp,\mathbf{D})=\mathit{W}(\mathbf{T}_g(\bg,\mathbf{D}),\mathbf{J}(\bs),\bp,\mathbf{W}_g(\bg,\bs)),\quad
\mathbf{T}_g(\bg,\mathbf{D})=\mathbf{G}+\mathbf{B}_g\bg+\mathbf{D}.
\end{equation}
For each garment category, a T-posed template mesh $\mathbf{G}$ is defined. On top of the base mesh, we add linear displacement deformation $\mathbf{B}_g$ controlled by PCA coefficients $\bg\in{\mathbb{R}^{64}}$. This low dimensional representation is effective in capturing size variations of a specific garment category under T-pose. To deform garments with dressed SMPL body, we share garment pose parameter $\bp$ with SMPL and use SMPL's skeleton $\mathbf{J}(\bs)$ as the binding skeleton of the garment. Instead of directly using the skinning weights of SMPL, a neural network is utilized to estimate the skinning weights $\mathbf{W}_g$ of the garment. \edit{This design makes garment mesh independent with SMPL mesh and makes our garment model can support more garment topology than SMPL+D methods~\cite{bhatnagar2019multi,pavlakos2018learning,alldieck2018video}, if providing corresponding garment training data.} To capture variations caused by different pose and interaction between clothing and body, we add a high-frequency displacement $\mathbf{D}\in{\mathbb{R}^{3|\mathcal{V}_g|}}$ for vertices of the clothing. \edit{In this paper, for the conciseness of writing symbol, we denote the displacement directly as $\mathbf{D}$ instead of a function of latent dependent variables, such as $\bg, \bp$.}

\subsection{Image to Dressed Body}
Given a near-front view RGB image depicting a posed subject dressed on specific garments, our model estimates its body shape, pose parameters and global translation with $\hat{\bs}\in{\mathbb{R}^{10}},\hat{\bp}\in{\mathbb{R}^{72}},\hat{\mathbf{t}}\in{\mathbb{R}^{3}}$ and the garment parameters $\hat{\bg}\in{\mathbb{R}^{64}}$ and $\hat{\mathbf{D}}$. Our model mainly consists of four modules: image encoder, classification module, skinning weight network, and displacement network. Fig.~\ref{fig:pipeline} shows our algorithm pipeline, and we will discuss the details of the last two modules.

\edit{Our image encoder uses the feature extraction of ResNet-18~\cite{he2016deep} and average pooling the final feature map to $8\times8$ size. From the map, a fully connected layer is used to get the latent feature.} Then, four fully connected layers are used to predict shape parameters $\hat{\bs}$, pose parameters $\hat{\bp}$, translation $\hat{\mathbf{t}}$ and shared garment latent feature $\mathbf{l}\in{\mathbb{R}^{256}}$. For pose parameters, instead of directly predicting the axis-angle representation parameters $\hat{\bp}$, we predict vectorized rotation matrices $\mathit{R}\hat{(\bp)}\in{\mathbb{R}^{24\times9}}$ of all joints, where $\mathit{R}$ is the Rodrigues rotation transformation. This strategy makes training more stable and continuous~\cite{lassner2017unite,pavlakos2019expressive,pavlakos2018learning}. 

From shared garment latent $\mathbf{l}$, two fully connected layers are used to predict upper and lower garment classify scores $\hat{\mathbf{u}}\in{\mathbb{R}^2}$ and $\hat{\mathbf{d}}\in{\mathbb{R}^4}$ separately. Then, we concatenate  $\hat{\bs}$ and $\mathbf{l}$ as input of a two-layer Multi-layer perceptron(MLP)~\cite{qi2017pointnet} to predict neutral garment shape parameters $\hat{\bg}$. After that, utilizing skinning weight and displacement networks, we get garment skinning weights $\hat{\mathbf{W}_g}$ and high-frequency displacements $\hat{\mathbf{D}}$, respectively. Finally, with predicted pose parameters, we can reconstruct the body shapes and dressed garments together.

\subsection{Skinning Weight Network}
\label{sec:SkinNet}
It is an open problem to estimate skinning weights for an arbitrary character given a binding skeleton hierarchy. Recently, Liu \etal~\cite{liu2019neuroskinning} proposed the first generic network to infer the skinning weights of various characters binding to the mutative skeleton hierarchy. Inspired by ~\cite{liu2019neuroskinning}, \edit{we design our skinning weight network to infer weights for neutral garments, and the network makes weights computation fast, differentiable and garment type independent.}

Our network predicts the skinning weights of a specific neutral garment $\mathbf{T}_g(\hat{\bg},\mathbf{0})$ binding to the skeleton $\mathbf{J}$ of corresponding neutral SMPL body $\mathbf{T}_b(\hat{\bs},\mathbf{0})$. We compute all distances of each vertex of $\mathbf{T}_g(\hat{\bg},\mathbf{0})$ to each joint point of $\mathbf{J}$. Then, the coordinate, normal, and distances of each vertex of $\mathbf{T}_g(\hat{\bg},\mathbf{0})$ are concatenated as the input feature for the network, and it computes the weights for all vertices. Our network uses MLP to change the vertex feature dimension and utilizes standard Residual Block~\cite{he2016deep} to extract features. Besides, we use graph convolution to aggregate the neighborhood information. In order to make our network applicable to different garment categories, we use GAT~\cite{velivckovic2017graph} graph convolution, whose filter weight learning is independent of mesh connectivity, and the weight is determined by the input feature on vertices only. This characteristic makes our network based on GAT suitable for different garment types. The architecture details can be found in the supplementary.

\subsection{Displacement Network}
The shape structure of the garment can be well reconstructed based on the PCA coefficients $\bg$. However, high-frequency details, such as folds caused by different pose, are beyond the representation ability of the linear model. We train a displacement network to regress the displacement of each garment vertex on top of the base mesh. 
For the displacement, we use a similar network structure with the skinning weight network. To improve the regression ability, we train an independent network for each garment category rather than a general network for all types. Moreover, we use spiral graph convolution~\cite{bouritsas2019neural} for each garment category, which has state-of-the-art regression ability for meshes with the same connectivity. To capture high frequency information, we project each vertex of deformed base garment $\mathbf{M}_g(\hat{\bg},\hat{\bs},\hat{\bp},\mathbf{0})$ on the image, and crop the $32\times32$ patch centered on the projected vertex. \edit{Then, for each vertex, we use a shared MLP to encode its patch into a latent feature, and concatenate the feature with shared garment latent $\mathbf{l}$, predicted SMPL shape parameter $\hat{\bs}$, garment shape parameter $\hat{\bg}$ as well as its coordinate, normal and skinning transformation together as its input feature for the displacement network.} The details of the neural network are given in the supplementary.


\subsection{Loss Function}

With our constructed  dataset, ground truth shape and pose parameters are available for all training data, thus it is natural to adopt supervised training. In this part, we denote predicted $\mathbf{M}_g(\hat{\bg},\hat{\bs},\hat{\bp},\hat{\mathbf{D}})$ and $\mathbf{M}_b(\hat{\bs},\hat{\bp}))$ as $\hat{\mathbf{M}}_g$ and $\hat{\mathbf{M}}_b$ separately. In the following, we will give the details on how to design the loss terms.

\textbf{Losses on shape parameters.} We directly adopt the MSE between predicted and ground truth shape parameters. The loss for SMPL body parameters and garment parameters are separately defined as:
\begin{equation}
\begin{aligned}
L_{Bp} = \|\hat{\bs}-\bs\|_2^2+\| \hat{\mathit{R}(\bp)} -\mathit{R}(\bp) \|_2^2+\|\hat{\mathbf{t}}-\mathbf{t} \|_2^2,\quad
L_{Gp} = \|\hat{\bg}-\bg\|_2^2.
\end{aligned}
\end{equation}

\textbf{Losses on geometry.} We supervise reconstructed geometries and joints with ground truth data. $\mathit{J}_B$ is the mapping to output posed 3D joints of SMPL body $\mathbf{M}_b$.
\begin{itemize}
\item Losses on reconstructed garment geometry and reconstructed body joints are separately defined as:
\begin{equation}
\begin{aligned}
L_{G} = \| \hat{\mathbf{M}}_g-\mathbf{M}_g \|_2^2,\quad
L_{J3D} = \| \mathit{J}_B(\hat{\mathbf{M}}_b)-\mathit{J}_B(\mathbf{M}_b) \|_2^2.
\end{aligned}
\end{equation}

\item Losses on displacements $\mathbf{D}$. To improve detail reconstruction ability, we use $\ell_{1}$ loss for each vertex of $\mathbf{D}$ and $\ell_{2}$ loss on laplacian coordinates of $\mathbf{D}$. $\mathcal{L}$ represents the laplacian coordinates mapping from a 3D mesh.
\begin{equation}
\begin{aligned}
L_{D1}= |\hat{\mathbf{D}}-\mathbf{D}|, \quad
L_{D2}= \|\mathcal{L}(\hat{\mathbf{D}})-\mathcal{L}(\mathbf{D})\|_2^2.
\end{aligned}
\end{equation}

\end{itemize}

\textbf{Losses of projection.} We use $\Pi$ to represent the camera projection of 3D geometries. All our training data share a common camera intrinsic matrix. The loss of body projections and garment projections are separately defined as:
\begin{equation}
\begin{aligned}
L_{B2D} = \| \Pi(\hat{\mathbf{M}}_b) - \Pi(\mathbf{M}_b) \|_2^2,\quad
L_{G2D} = \| \Pi(\hat{\mathbf{M}}_g) - \Pi(\mathbf{M}_g) \|_2^2.
\end{aligned}
\end{equation}

\textbf{Losses of classification.} We use standard softmax loss to penalize the classification error of $\hat{\mathbf{u}}$ and $\hat{\mathbf{d}}$ relative to ground truth garment types.

\textbf{Losses of interpenetration.} During training, inferred garments and body are easy to occur interpenetration. We use a simple yet effective interpenetration term inspired by ~\cite{gundogdu2019garnet} to alleviate this problem:
\begin{equation}
L_{int}(\mathbf{P},\mathbf{Q})=\!\!\!\!\!\!\!\!\sum_{\{i,j\}\in\mathcal{C}(\mathbf{P},\mathbf{Q})}\!\!\!\!\!\!\!\!ReLU(-\mathbf{n}_{\mathbf{q}_j}^T(\mathbf{p}_i-\mathbf{q}_j))/N,
\label{eq:int}
\end{equation}
where $\mathbf{P}$,$\mathbf{Q}$ are two interpenetrated meshes. $\mathcal{C}(\mathbf{P},\mathbf{Q})$ represents the valid corresponding pairs between $\mathbf{P}$ and $\mathbf{Q}$, and these pairs are filtered based on distances and normal angles. This loss penalizes vertex $\mathbf{p}_i$ that is inside the local plane defined by its corresponding point $\mathbf{q}_j$ and its normal $\mathbf{n}_{\mathbf{q}_j}$. We use this loss on reconstructed neutral garments and body as well as posed garments and body separately:
\begin{equation}
\begin{aligned}
L_{inters}=L_{int}(\mathit{T}_g(\hat{\bg},\mathbf{0}),\mathit{T}_b(\hat{\bs},\mathbf{0}))+L_{int}(\hat{\mathbf{M}}_g,\hat{\mathbf{M}}_b)
\end{aligned}
\end{equation}

\textbf{Loss of Skinning Weight Network.} As discussed in~\cite{liu2019neuroskinning}, the weight vector $\{\omega_{ij}|j\in{|\mathit{J}(\bs)|}\}$ of $\mathbf{W}_g$ is a selection of different bones with different probabilities. We use the Kullback-Leibler divergence loss to measure the distance between predicted weights distribution $\hat{\omega_{ij}}$ and ground truth distribution $\omega_{ij}$:
\begin{equation}
\label{equ:ws_loss}
L_{ws}=\sum_{i=1}^{|\mathcal{V}_g|}\sum_{j=1}^{24}\hat{\omega}_{ij}(\log \frac{\hat{\omega}_{ij}}{\omega_{ij}}).
\end{equation}

To train the whole network, we first train the skinning weight network with loss in Eq.~\eqref{equ:ws_loss}, and then train other parts together by fixing the skinning weight network.
\section{Dataset Construction}


\subsection{Skinning Weight Dataset}
To train our skinning weight network, we need some neutral garments with ground truth skinning weights. \edit{Our network training adapts to any weight calculation method. For simpleness, we compute garment weights from the dressed SMPL body.}

For vertex $\mathbf{p}_i$ of the garment, we select $K$ vertices from dressed body mesh, based on distance, normal angle, and segmentation prior. Segmentation prior is some information we can utilize, such as corresponding vertices of right trouser leg must belong to the right leg of body mesh. From selected $K$ vertices of body, we average their skinning weights with IDW(inverse distance weighting) as the skinning weight of $\mathbf{p}_i$. After all vertices' weights have been computed, we apply Laplacian smoothing~\cite{Sorkine05} to remove noises and artifacts.

With this method, for all garment types, we construct a skinning weights dataset, which includes $48000$ neutral garments for training, and $6467$ for test.

\subsection{Synthetic Dataset Construction}

As there does not exist publicly available dataset containing pairs of the color image and corresponding body and cloth shapes, we construct the dataset with a physics-based simulation method. The dataset construction process can be divided into four steps: sewing pattern design, neutral garment synthesis, posed garment simulation and rendering. ~\cite{wang2018learning} proposed a novel method to synthesize neutral garments. We extend their method to support more garment types and posed garments generation.

\begin{figure}
\begin{center}
\includegraphics[width=\linewidth]{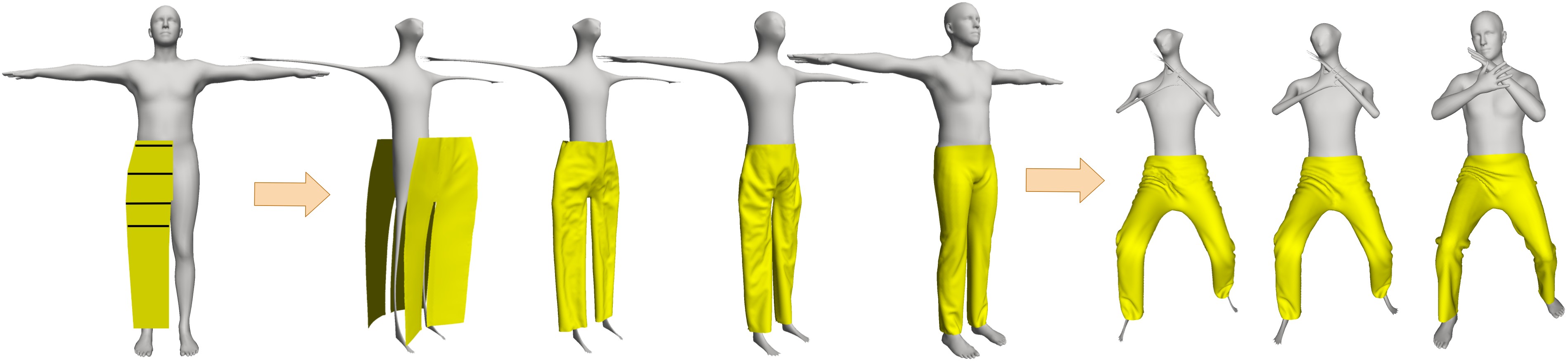}
\end{center}
   \caption{Our synthesis process of a pant. First, we generate a random sewing pattern based on neutral body type. Then, we stitch the pattern on the skeleton and inflate the skeleton to its original shape to generate the neutral pant. Finally, we skinning deform the skeleton and neutral pant to a posture, and simulate the final pant with gravity by inflating the posed skeleton.
   }
\label{fig:pant_gene}
\end{figure}

As shown in Fig.~\ref{fig:pant_gene}, we first design the pant sewing pattern based on body type. Then, around the neutral skeleton, we connect the sewing lines of the front and back pattern and shorten the length gradually. The sewing lines are stitched together after all lengths of the sewing line are less than a threshold. To simulate the realistic result of the garment draped on the neutral body, we inflate the skeleton and add gravity. For posed garment simulation, we deform the neutral garment to target pose with generated skinning weights and inflate the body and add gravity to simulate the posed garment. In this work, we assume that both the human body and the garment are in a statically stable state. Therefore, we sample discrete pose instead of simulating the whole motion sequence.

After generating the garment shapes, the synthetic images are rendered by following the methods in~\cite{liu2019temporally,liang2019shape,varol2017learning}. By randomly selecting body textures from SURREAL~\cite{varol2017learning}, garment textures from Fabrics~\cite{kampouris2016fine} and DTD~\cite{cimpoi2014describing}, background images from Places365-Standard dataset~\cite{zhou2017places} and global illumination from hundreds HDR images, we can render near-front view dressed body images with abundant variations.

\begin{figure}
\begin{center}
\includegraphics[width=\linewidth]{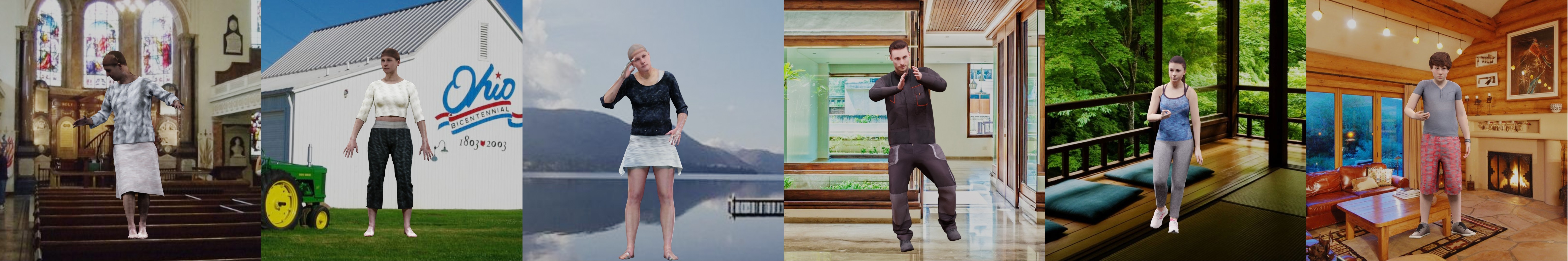}
\end{center}
   \caption{Some examples from our synthetic dataset(left three) and HD texture dataset(right three).}
\label{fig:eg_imgs}
\end{figure}

We implement the abovementioned pipeline using the simulator NvFlex\cite{NvFlex} and Blender\cite{Blender}. We utilize 3048 body shape of SPRING dataset~\cite{yang2014semantic}, and randomly generate neutral clothes dressed on them. For posed garment, we select 55 motion sequences from CMU Mocap~\cite{Mocap}, whose poses have been converted to SMPL pose parameters with MoSh~\cite{loper2014mosh}. For each motion sequence, we randomly select 10 different persons with 4 sets of different clothing separately and sample pose parameters every 30 frames. Finally, we get 168602 dressed bodies as training data and 8874 as test data. The left part of Fig.~\ref{fig:eg_imgs} shows several examples of our synthetic images.

\begin{figure}
\begin{center}
\includegraphics[width=\linewidth]{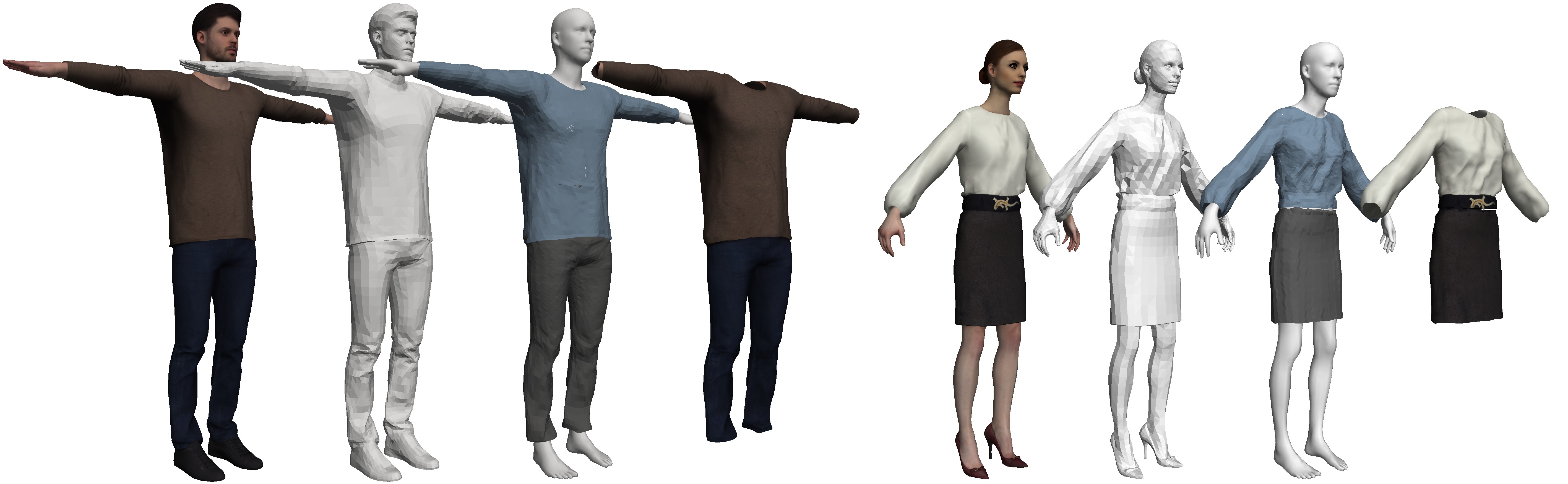}
\end{center}
   \caption{Two examples of rigged avatar registration. We show the scanned meshes with and without texture, reconstructed geometries and garment with texture in each group.}
\label{fig:rig_rec}
\end{figure}

\subsection{HD Texture Dataset}
Although synthetic samples are visually realistic, they still have a noticeable domain difference with real images. Therefore, we process another dataset with high-definition (HD) textures. We purchase 104 and 181 rigged avatar from RenderPeople~\cite{RP} and Axyz~\cite{AXYZ}, respectively. These avatars have high-quality geometry and realistic texture. We use Mixamo~\cite{Mixamo} to drive avatars and get about 89425 posed meshes as training data and 4386 as test data. The abovementioned rendering pipeline is used to produce high-quality images, and the right part of Fig.~\ref{fig:eg_imgs} shows some examples. Because the body and clothes part of the scanned mesh are not separated, and the connectivities of scanned meshes are not consistent, we need to process these meshes to our representation via the following two steps.

\textbf{Rigged registration.} For a rigged mesh with A- or T-pose, we segment it to garment and skin parts. We optimize garment shape parameters $\bg$, displacements $\mathbf{D}$, body parameters $\bs$, $\bp$ and translation $\mathbf{t}$ to register our representation to the avatar. We penalize the point-to-plane distance for both reconstructed garment and body. And we use Eq.~\eqref{eq:int} to reduce the interpenetration among them. To get a size matching garment, we adopt the rendered silhouette loss utilizing~\cite{kato2018neural}. And we add $\ell_{2}$ regular term for garment and body parameters. With this pipeline, we reconstruct all garments and body shapes of rigged avatars, and we extract texture for each garment. Fig.~\ref{fig:rig_rec} shows two examples.

\textbf{Posed registration.} After we finish the rigged avatar reconstruction, we initialize our posed model optimization with rigged reconstruction parameters and optimize pose parameters $\bp$ and translation $\mathbf{t}$ first. And then, we fine-tune all parameters to get final posed reconstruct results.

\begin{figure}
\begin{center}
\includegraphics[width=\linewidth]{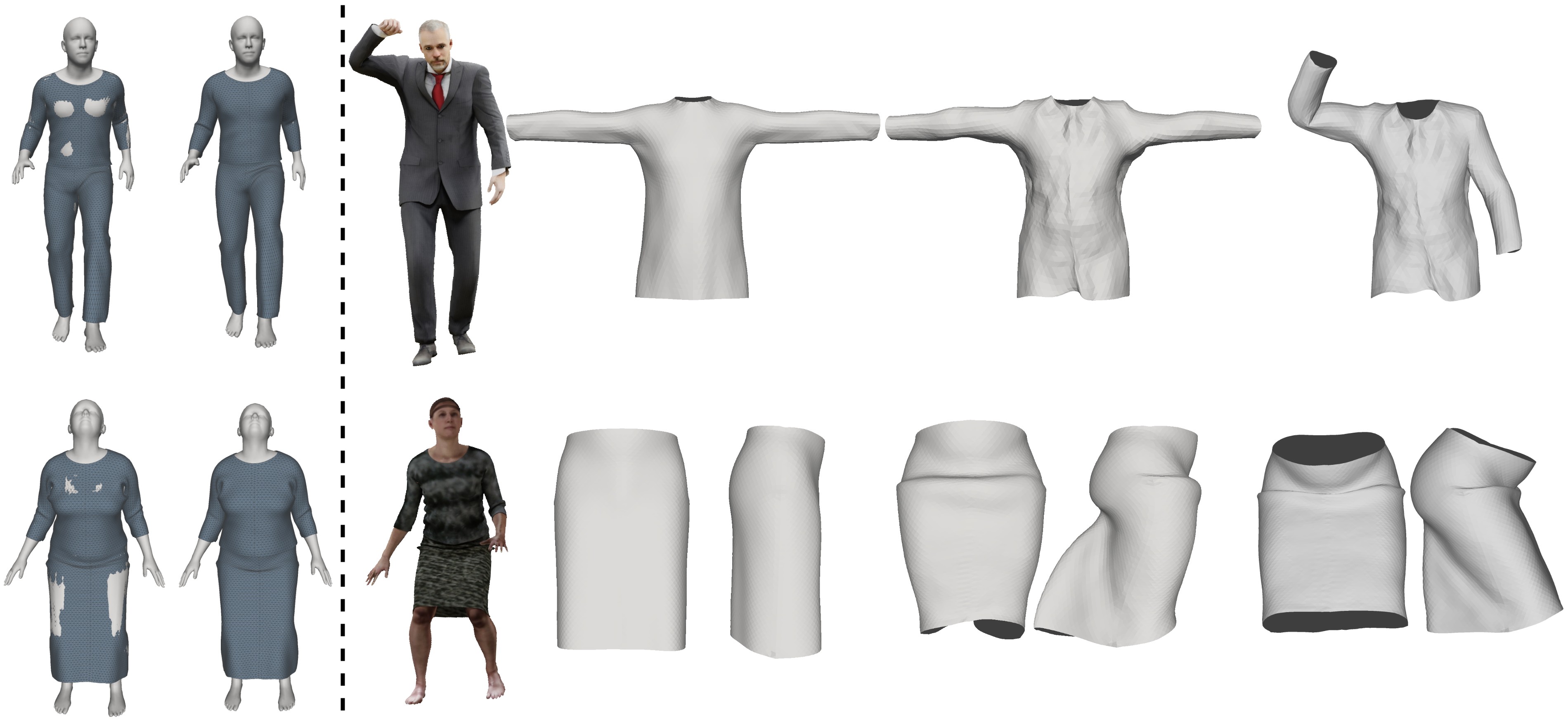}
\end{center}
   \caption{The left shows the ablation study for interpenetration loss. The examples demonstrate that the interpenetration term in Eq.~\eqref{eq:int} alleviates the collision problem. The right shows our predicted displacement results. \edit{For each example, We present the base mesh, mesh with displacement and reposed mesh, respectively.}  The first example captures detail geometry on top of the base mesh, and the second one recovers large scale deformation for the skirt caused by leg movement. For better visualization, we show two viewpoints for the second result.}
\label{fig:displacement}
\end{figure} 

\section{Experiments}
In this part, we first evaluate our BCNet. Then we quantitatively compare with state-of-the-art methods. Finally, we present some qualitative results. More results are supplied in the supplementary.

\subsection{Analysis of BCNet}
\textbf{Our Test Set.} We test our predicted errors on Synthetic and HD Texture test set, respectively. Table~\ref{tab:test_set} shows the mean Euclidean distance(MED) of reconstructed shapes after Procrustes transformation and ground truth shapes.

\textbf{Skinning Weight Network.} We test the reconstruction ability of the skinning weight network on our test dataset. For each garment, we reconstruct its skinning weights with our network. The average $\ell_{1}$ reconstruction error on the whole test set is $6.5\times10^{-4}$. Then, we sample $20$ poses from the Mocap dataset and deform the neutral clothes to the posture with our predicted weights and ground truth weights separately. The average MED of reposed mesh for all garment types is $0.43$mm. These results demonstrate that our skinning weight network can reach very high accuracy. More details are given in supplementary.

\begin{table}
\parbox{.4\linewidth}{
\centering
\caption{The MED(cm) between predicted and ground truth shapes on our test dataset. For garments, we report errors with(gray) and without(white) displacement module, respectively.}
\label{tab:test_set}
\begin{tabular}{|c|c|c|c|c|}
\hline
dataset &shirt & pant & skirt & body \\
\hline
\multirow{2}{4em}{\centering Synthetic} & \cellcolor{lightgray} 0.91 & \cellcolor{lightgray}0.75 & \cellcolor{lightgray}0.87 & \multirow{2}{2em}{\centering 1.57} \\
\cline{2-4}
& 1.72 & 1.59 & 2.46 & \\
\hline
\multirow{2}{4em}{\centering HD Texture} & \cellcolor{lightgray}1.71  & \cellcolor{lightgray}1.42 & \cellcolor{lightgray}1.65 & \multirow{2}{2em}{\centering 2.93}  \\
\cline{2-4}
 & 1.97  &1.72 & 1.87 &  \\
\hline
\end{tabular}

}
\parbox{.6\linewidth}{
\centering
\caption{The errors(cm) on BUFF rough A-pose dataset(gray) and Digital Wardrobe dataset(white).}
\label{tab:test_buff}
\begin{tabular}{|c|c|c|c|c|}
\hline
Methods &Upper & Lower & Total & Chamfer \\
\hline
\rowcolor{lightgray} MGN-opt-8 & 1.63 & 1.91 & 1.82 & 1.91 \\
\rowcolor{lightgray} MGN-8 & 1.78  & 2.13 & 1.99 & 2.08  \\
\rowcolor{lightgray} Octopus-opt-8 & 1.40  & \textbf{1.35} & \textbf{1.31} & 1.41  \\
\rowcolor{lightgray} Octopus-8 & 1.54  & 1.74 & 1.70 & 1.76  \\
\rowcolor{lightgray} Ours & \textbf{1.07}  & \textbf{1.35} & 1.35 & \textbf{1.34}  \\
\hline
PIFu & 1.59 & \textbf{1.37} & 1.85 & \textbf{1.61} \\
DeepHuman & 2.38  & 2.46 & 3.15 & 2.98  \\
Ours & \textbf{1.44}  & 1.78 & \textbf{1.80} & 1.77  \\
\hline
\end{tabular}
}
\end{table}

\textbf{Interpenetration.} Our network infers human body and layered garments mesh separately, which brings better flexibility but at the cost of introducing more complex interactions between body and garments. Interpenetration is a common unreal phenomenon which is very easy to perceive by a human. Therefore, it is quite necessary to process interpenetration between these meshes. We propose an interpenetration term in Eq.~\eqref{eq:int} to alleviate this problem, and an ablation study on this term is shown in the left of Fig.~\ref{fig:displacement}. We can see that the interpenetration loss is beneficial to alleviate the interpenetration problem.

\textbf{Displacement Network.} Garment PCA shape parameter $\bg$ can represent the garment structure, while it can not represent the detailed shape of a specific garment and large scale deformations caused by pose and gravity for loose garments. We train our non-linear displacement network to expand the representation ability. The result of ablation study on displacement network is given in Table~\ref{tab:test_set}, we can observe that the displacement network greatly improves the reconstruction accuracy. In the right part of Fig.~\ref{fig:displacement}, we show two examples of our displacement results. We present input image, base garment $\mathbf{T}_g(\bg, \mathbf{0})$, \edit{the garment with predicted displacement $\mathbf{T}_g(\bg,\mathbf{D})$ and final reposed garment $\mathbf{M}_g(\bg,\bs,\bp,\mathbf{D})$} for each result. In the first row, we show an example of predicted displacement capturing detailed geometry, such as tie and suit boundary line. In the second row, we show large scale deformations on a skirt caused by bending leg motion, and we use two viewpoints to show the deformation results.

\subsection{Quantitative Comparison.}

We test our reconstruction accuracy on two public data sets, BUFF~\cite{zhang2017detailed} and Digit Wardrobe(DW)~\cite{bhatnagar2019multi}. We segment the ground truth scan mesh into upper, lower garment and body parts, and compute error for garments and whole clothed bodies separately. Because our model predicts separate body and garment meshes, we extract the outer surface of all meshes as the proxy to do registration and error computing for a fair comparison. We measure the average point-to-surface Euclidean distance(P2S) in cm from the ground truth to predicted surface for upper, lower garments, and the whole surface. We also compute the Chamfer distance~\cite{saito2019pifu} between the reconstructed and the ground truth surfaces.

\begin{figure}
\begin{center}
\includegraphics[width=\linewidth]{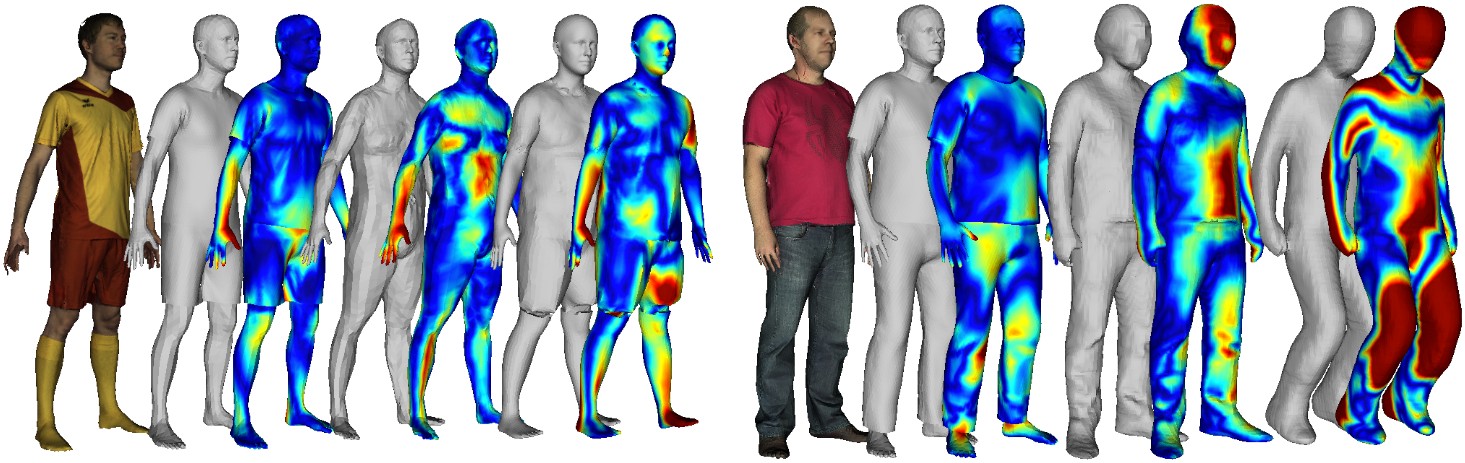}
\end{center}
   \caption{Error maps on BUFF(left part) and MGN(right part). From left to right, we show the GT mesh, results of ours, Octopus-opt-8, MGN-opt-8 for the BUFF example, and results of ours, PIFu, DeepHuman for the MGN example(red means $\geq4$cm). }
\label{fig:compare}
\end{figure}

\textbf{BUFF Dataset.} We compare the reconstruction accuracy of our method with SMPL+D based methods octopus~\cite{pavlakos2018learning} and MGN~\cite{bhatnagar2019multi}. By default, their methods require multi-view semantic segmentation images and 2D joints of a clothed body under rough A-pose as inputs, and post-optimization is applied to refine the results. Therefore, we select 21 rough A-pose data from BUFF~\cite{zhang2017detailed} as our test set. Table~\ref{tab:test_buff} shows our results, and their results of 8 perspective inputs with and without optimization, respectively. Although the input of our method only needs one image, our method can get better numerical results than theirs without post-optimization, and an equivalent result with Octopus with optimization. The post-optimization is time-consuming and takes several seconds and several minutes for Octopus and MGN, respectively. For MGN, we manually modified some segmentation error of PGN~\cite{gong2018instance} to refine their results. In the left part of Fig.~\ref{fig:compare}, we show an example of our result and their results with post-optimization. Some unnatural folds are introduced in the post-optimization step of MGN while our method does not have this problem.

\textbf{DW dataset.} Digital Wardrobe~\cite{bhatnagar2019multi} includes registered clothed body meshes with real texture under more general posture. We use 94 meshes to compare with non-parametric methods PIFu~\cite{saito2019pifu} and DeepHuman~\cite{zheng2019deephuman}\footnote{We did not test~\cite{pavlakos2018learning,bhatnagar2019multi} on this dataset as most of the samples are not A-pose.}. Table~\ref{tab:test_buff} shows the results. For PIFu with single image input, our method can get similar reconstruction accuracy. However, the reconstructed results by PIFu combine both shapes in one mesh without semantic information, while our method can fully control the predicted separate body and cloth meshes. The results of DeepHuman tend to bend the leg, which introduces large errors for this dataset. The right part of Fig.~\ref{fig:compare} shows an example of the results.

\begin{figure}
\begin{center}
\includegraphics[width=\linewidth]{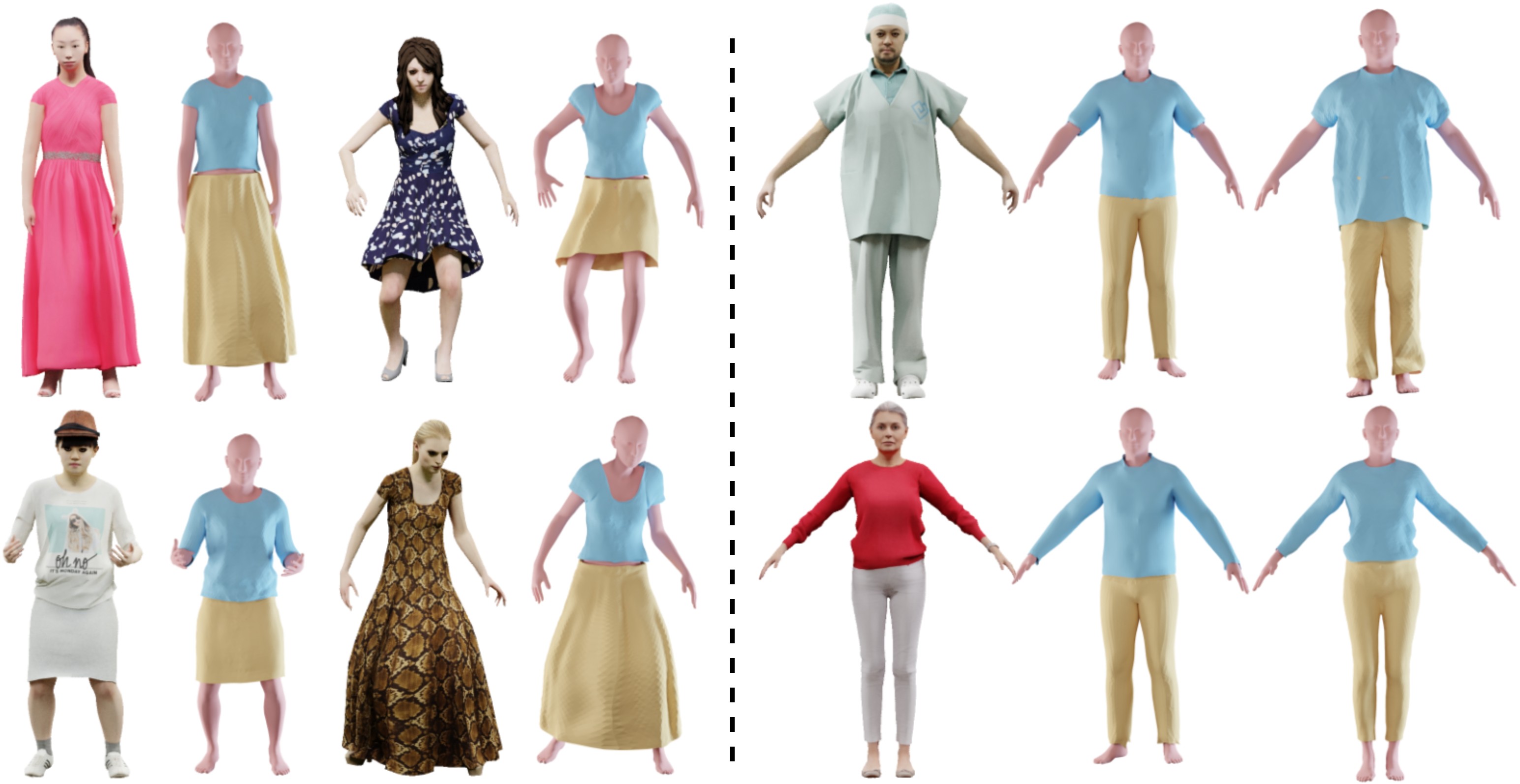}
\end{center}
   \caption{The left part: reconstructed body and garment shapes by our method on four images of our test set. The right part: qualitative comparison between \edit{the results of MGN~\cite{bhatnagar2019multi} without post-optimization} and ours. In each group, the input image, result of MGN, and ours are displayed respectively.}
\label{fig:quality_comp}
\end{figure}

\subsection{Qualitative Results.} 
In this following, we show some visual results of our method and the comparison with MGN. As our method can reconstruct the body and garments separately, garment transfer between two input images can be achieved. Some garment transfer results are given in the supplementary.

\textbf{Reconstruction Quality.} In the left part of Fig.~\ref{fig:quality_comp}, we present our reconstructed body and garments shapes on several test images. Our method can recover accurate body posture and capture the garment geometry to some extent from a single input image. Thanks to our separated garment representation with adaptive skinning weights, we can reconstruct plausible shape for loose garments with large edges.

\textbf{Comparison with MGN.} As a template-based method, MGN~\cite{bhatnagar2019multi} is the most relevant prior method with ours. MGN represents garment by binding the garment to SMPL vertices and uses a mask to select valid vertices for a specific garment type. MGN needs multi semantic segmentation images as input and constrains the posture to rough A-pose. Besides, MGN needs a time-consuming post-optimization step to refine the predicted result. Differently, Our method only requires one frontal view image with arbitrary posture and directly produces the final results from the network. In the right part of Fig.~\ref{fig:quality_comp}, we show two qualitative comparisons with MGN. Our method can generate more accurate body shape and size of garments, while the results of MGN without post-optimization have similar shapes for different inputs and lack garment details.

\section{Conclusion}
We introduced BCNet, a novel method to automatically reconstruct both body and garment shapes from a single RGB image. Rather than binding garment with SMPL like prior SMPL+D based representation, our proposed model can produce layered garments with different topology and skinning weights, which makes BCNet a model capable of jointly reconstructing body and loose garment, like skirts. To train BCNet, we designed a complete algorithm pipeline to generate clothed body data. Experiments demonstrated that our method can generate comparable or better reconstruction results compared with state-of-the-art methods, while allowing more flexible controls such as garment transfer. Our constructed dataset and our proposed BCNet would push a step for the research on digitizing human.

\section*{Acknowledgments}
This work was supported by the National Natural Science Foundation of China (No. 61672481), and Youth Innovation Promotion Association CAS (No.2018495).
%
%
\bibliographystyle{splncs04}
\bibliography{egbib}
\appendix
\section*{Appendix}
In the main paper, we propose the BCNet, a deep neural network model that takes a near front view color image of the clothed human body as input, and outputs reliable garments and body 3D geometries separately. In this supplemental material, we first describe some \edit{implementation details}, then present more qualitative results and discuss the limitations of our approach.

\begin{figure}
\begin{center}
\includegraphics[width=\linewidth]{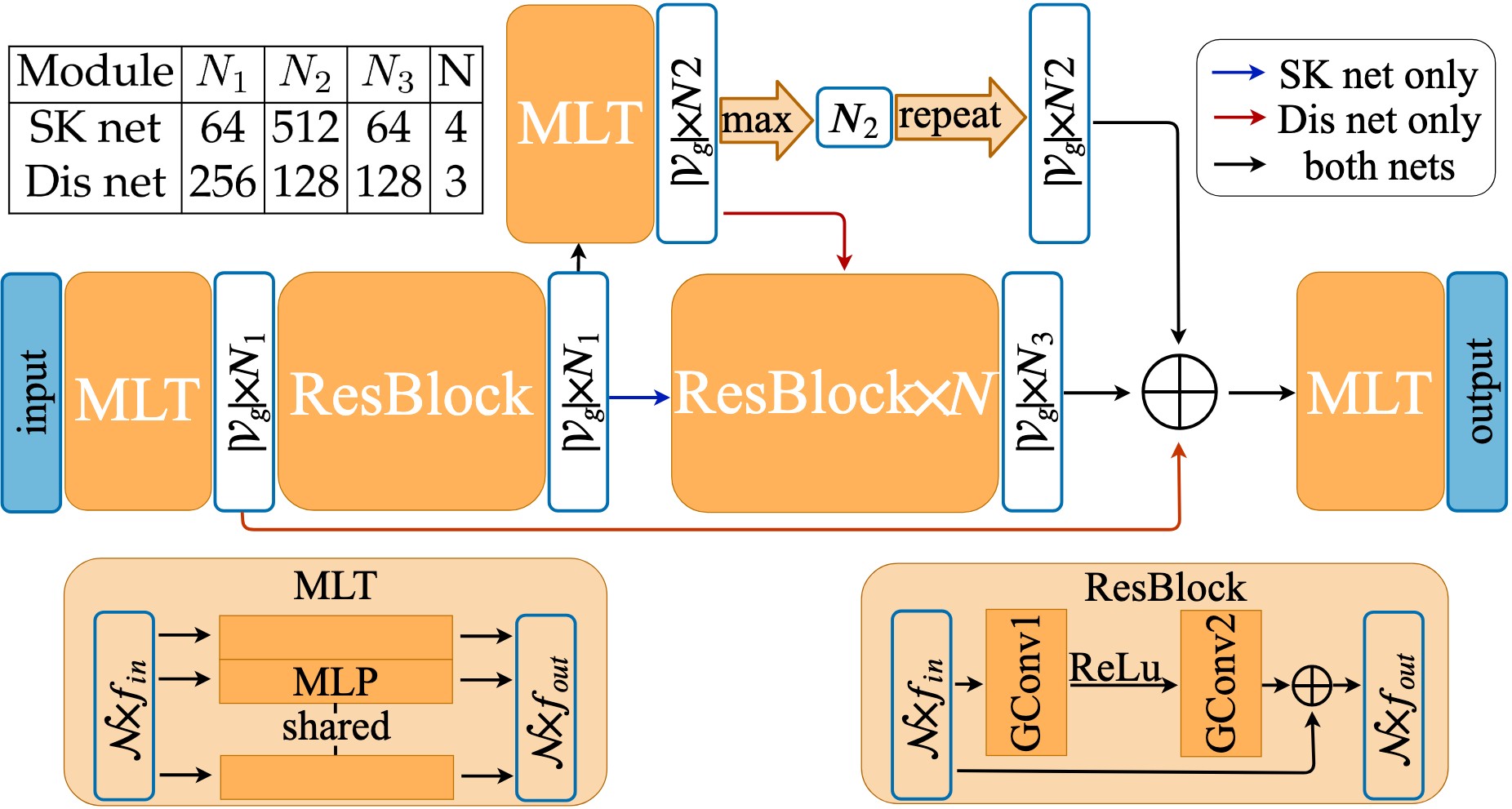}
\end{center}
   \caption{The architecture of the skinning weight(SK net) and displacement network(Dis net). Two networks follow similar architecture with slightly different configuration, as presented in the figure. The Multi-Layer transform(MLT) is a module that utilizes a shared Multi-Layer perceptron(MLP) to change the feature dimension for each vertex. The ResBlock follows the design of the standard residual network.}
\label{fig:architecture}
\end{figure}

\section{Implementation Details}

\textbf{Detailed Architecture.} In this section, we describe the detailed architecture design of the skinning weight network and displacement network. Both networks follow similar architecture presented in Fig.~\ref{fig:architecture} while with slightly different module configuration and computing flow. The architecture is a point-to-point feature calculation process that does not require upsampling or downsampling. We utilize Multi-Layer transform(MLT) and ResBlock to construct the network. The MLT uses a shared Multi-Layer perceptron(MLP) to change the feature dimension for each vertex. And the ResBlock is used to extract deeper features following the classical design. To obtain global information, we do max-pooling on the middle layer feature and concatenate it to the input of the final MLT. Optionally, we can also concatenate the shallow feature to enhance the inference ability of high-frequency details. The specific graph convolution, input features, and outputs for the two networks are different, and they have been described in the main paper.

\edit{\textbf{Details of Rigged and Posed Registration.} Before computation, we need to segment the rigged avatars. Some of the purchased rigged avatars provide accurate garment and skin segmentation. For segmentation that is inaccurate or unavailable, we manually segment the models with Blender utilizing the texture or color information. For BUFF[56] data, we segment similarly. The garment and skin segmentation of the Digital Wardrobe [7] is available. For the optimization, we implement the whole objective energy with Pytorch framework, and minimize the loss with the Adam iteration method.}

\section{More Results}
\begin{figure}[t!]
\begin{center}
\includegraphics[width=\linewidth]{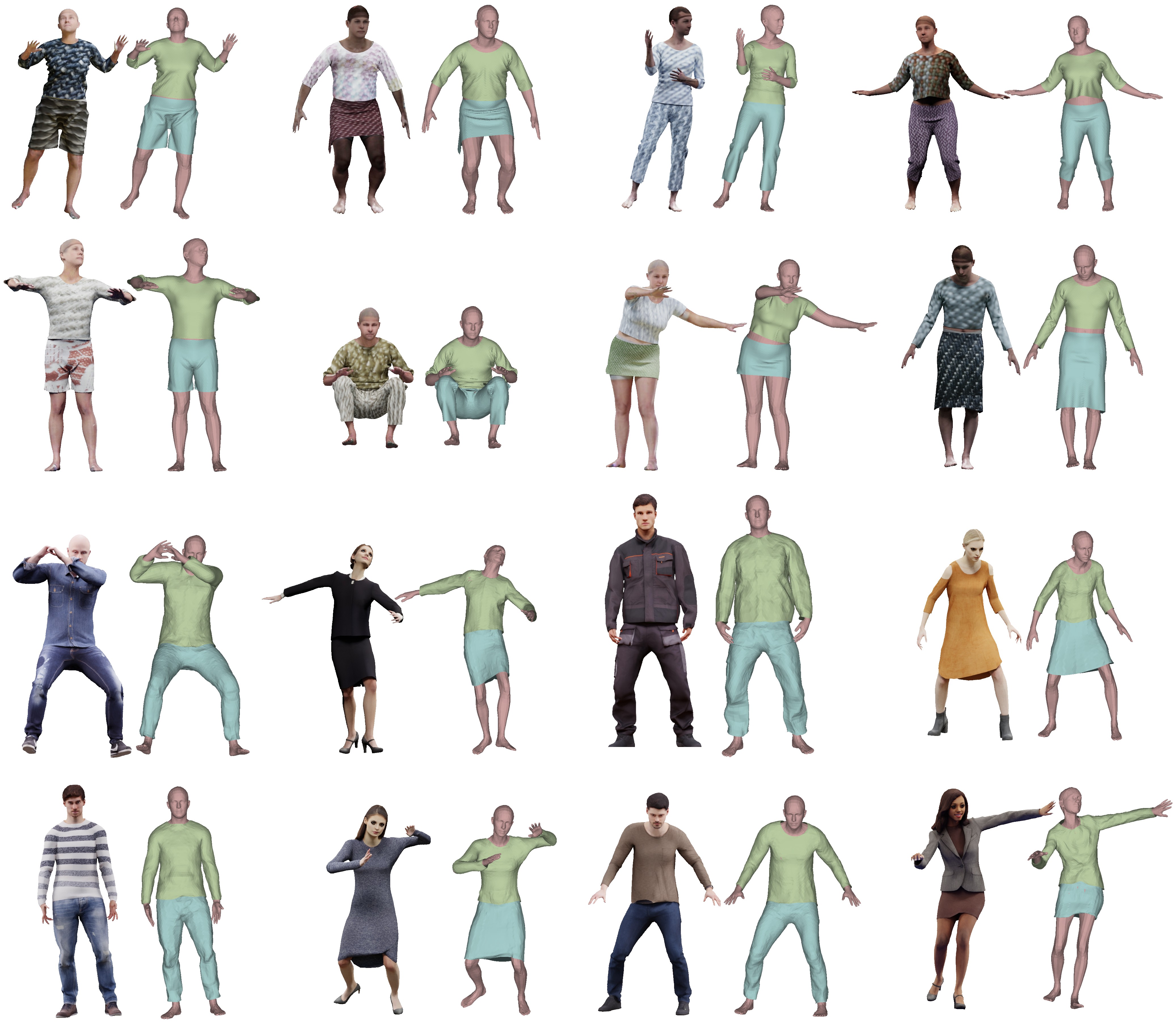}
\end{center}
    \caption{Some examples of Synthetic Dataset and HD Texture Dataset. We show color image and geometry in each group. The first two rows are samples from Synthetic Dataset, and the last two rows are samples from HD Texture Dataset. Our constructed dataset includes various kinds of postures and garments.}
\label{fig:gt_data}
\end{figure}

\begin{figure}[htb]
\begin{center}
\includegraphics[width=0.9\linewidth]{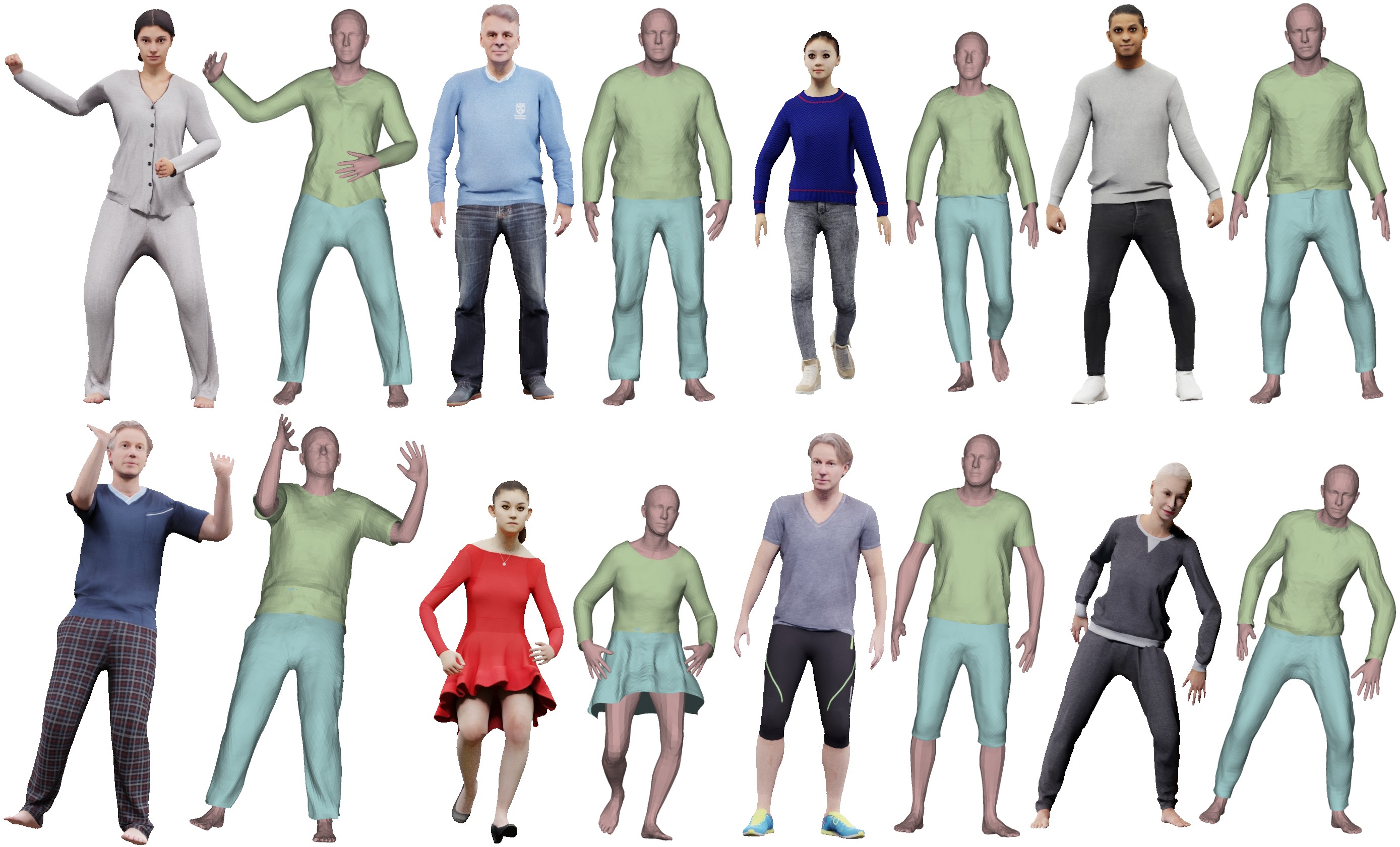}
\end{center}
   \caption{Reconstructed results by our method on test set. Each group includes the input image and our result.}
\label{fig:HD_test}
\end{figure}

\begin{figure}[htb]
\begin{center}
\includegraphics[width=0.92\linewidth]{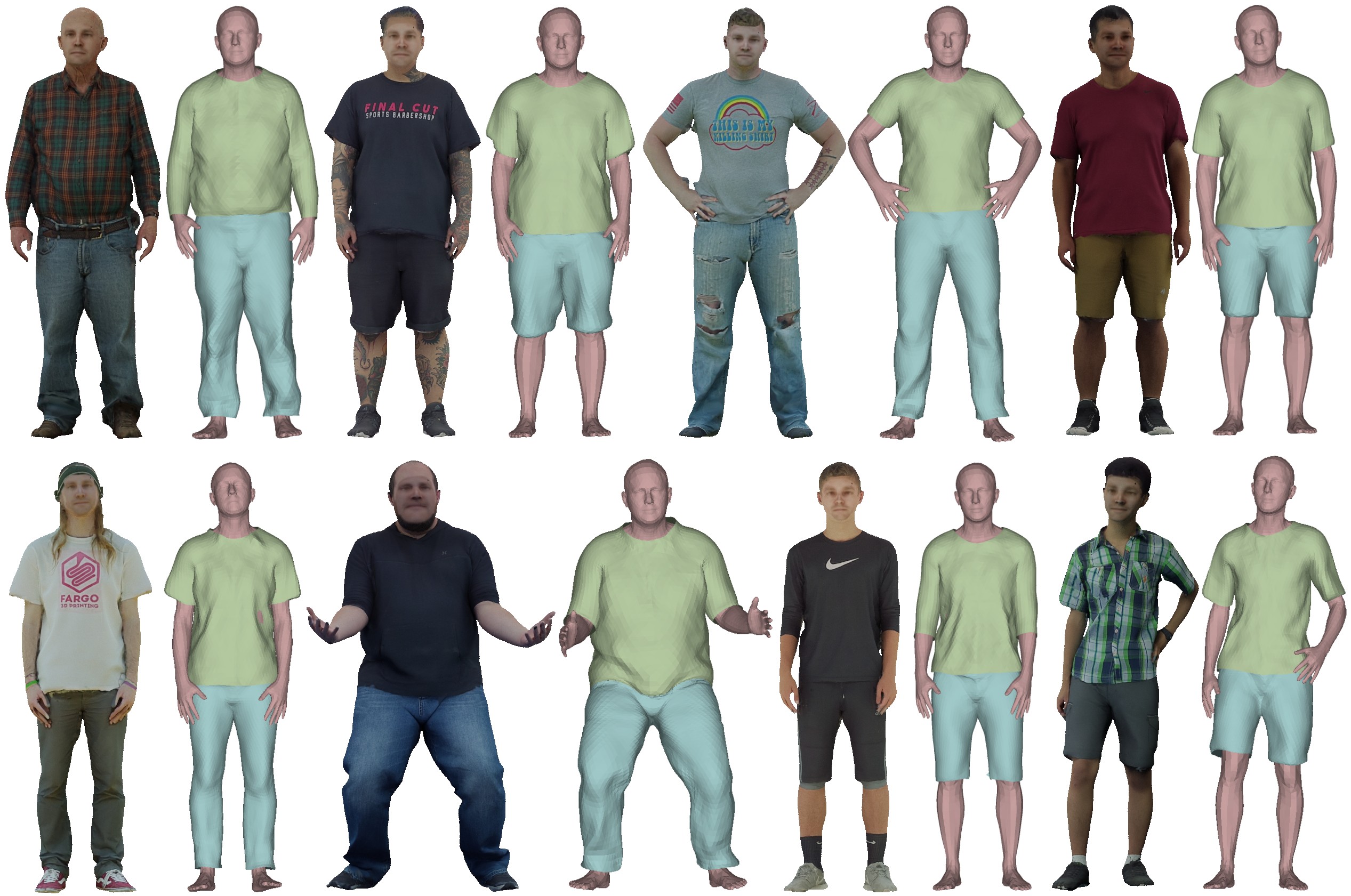}
\end{center}
   \caption{Reconstructed results by our method on DW Dataset~[7]. Our method can recover the body shape, posture, and cloth type and shape quite well.}
\label{fig:DW}
\end{figure}

\begin{figure}[htb]
\begin{center}
\includegraphics[width=0.9\linewidth]{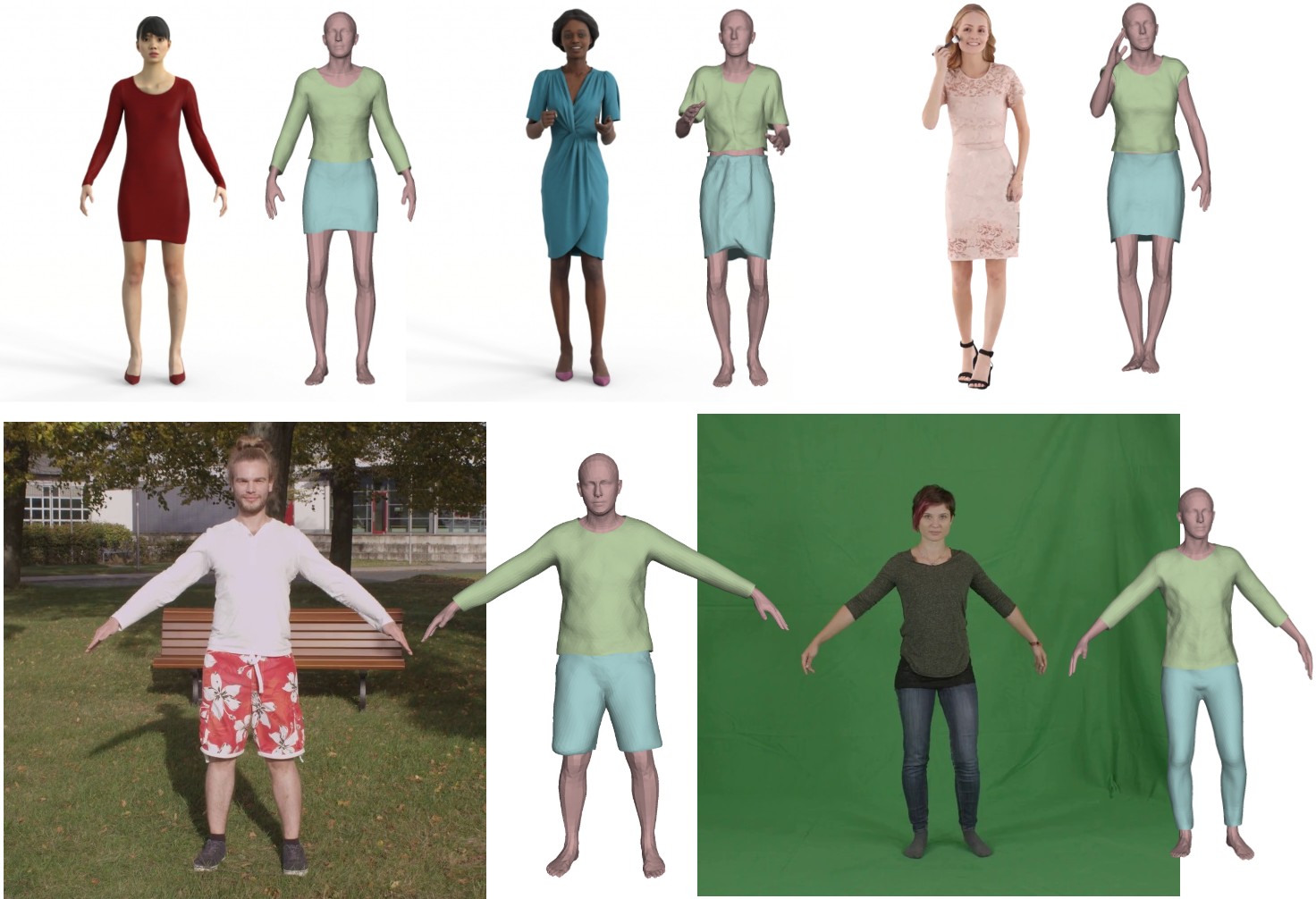}
\end{center}
   \caption{Reconstructed results by our method on real images. Each group includes the input image and our result. First row is three images from the internet, and second row is two images from PeopleSnapshot~[3] dataset.}
\label{fig:real}
\end{figure}

\begin{figure}[t!]
\begin{center}
\includegraphics[width=\linewidth]{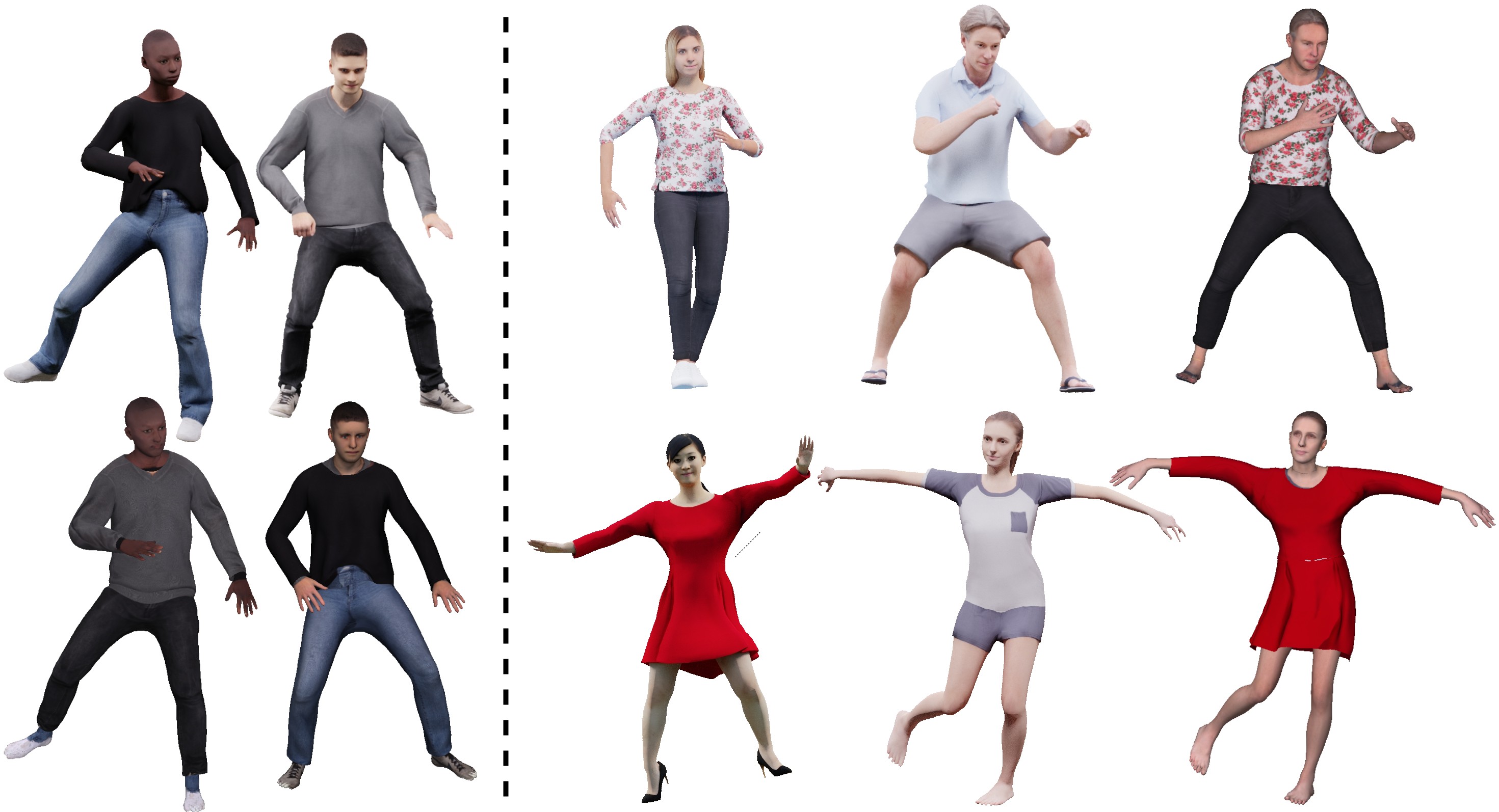}
\end{center}
   \caption{On the left, with two input images of the first row, we can predict their body shapes and exchange garments geometries and textures with BCNet. In the right, two garment transfer examples are given. In each row, we transfer the garments of the first image to the second image and show the reconstructed shapes with texture on the third column. }
\label{fig:exchange}
\end{figure}

\begin{figure}[t!]
\begin{center}
\includegraphics[width=\linewidth]{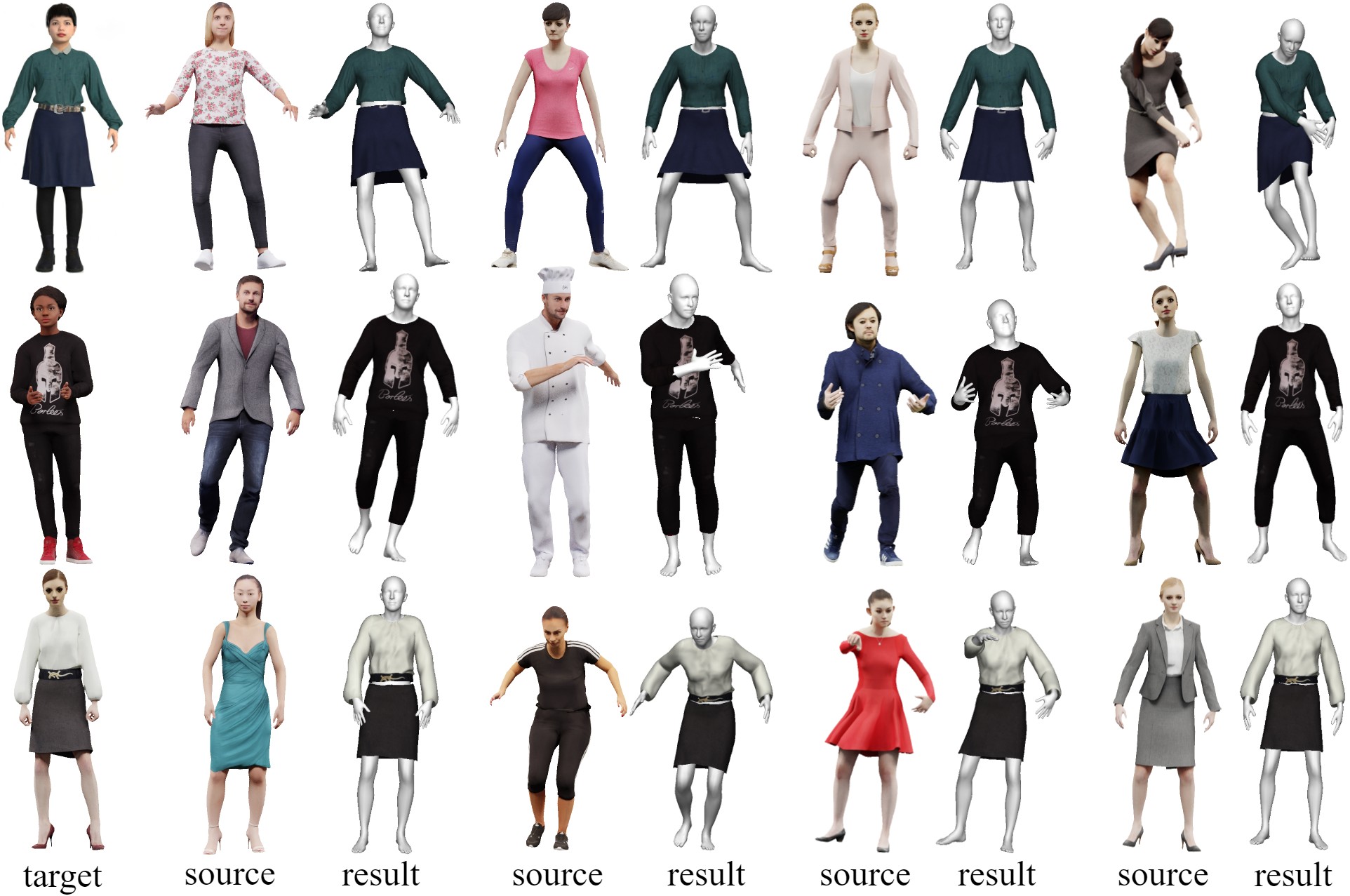}
\end{center}
    \caption{More garment transfer results on test dataset. The target garment images are shown in the first column, and the garment in the target image is transfered to four different source clothed body images. For each result, the predicted body and transfered garments with texture are presented.}
\label{fig:gar_transfer}
\end{figure}

\begin{figure}[t!]
\begin{center}
\includegraphics[width=\linewidth]{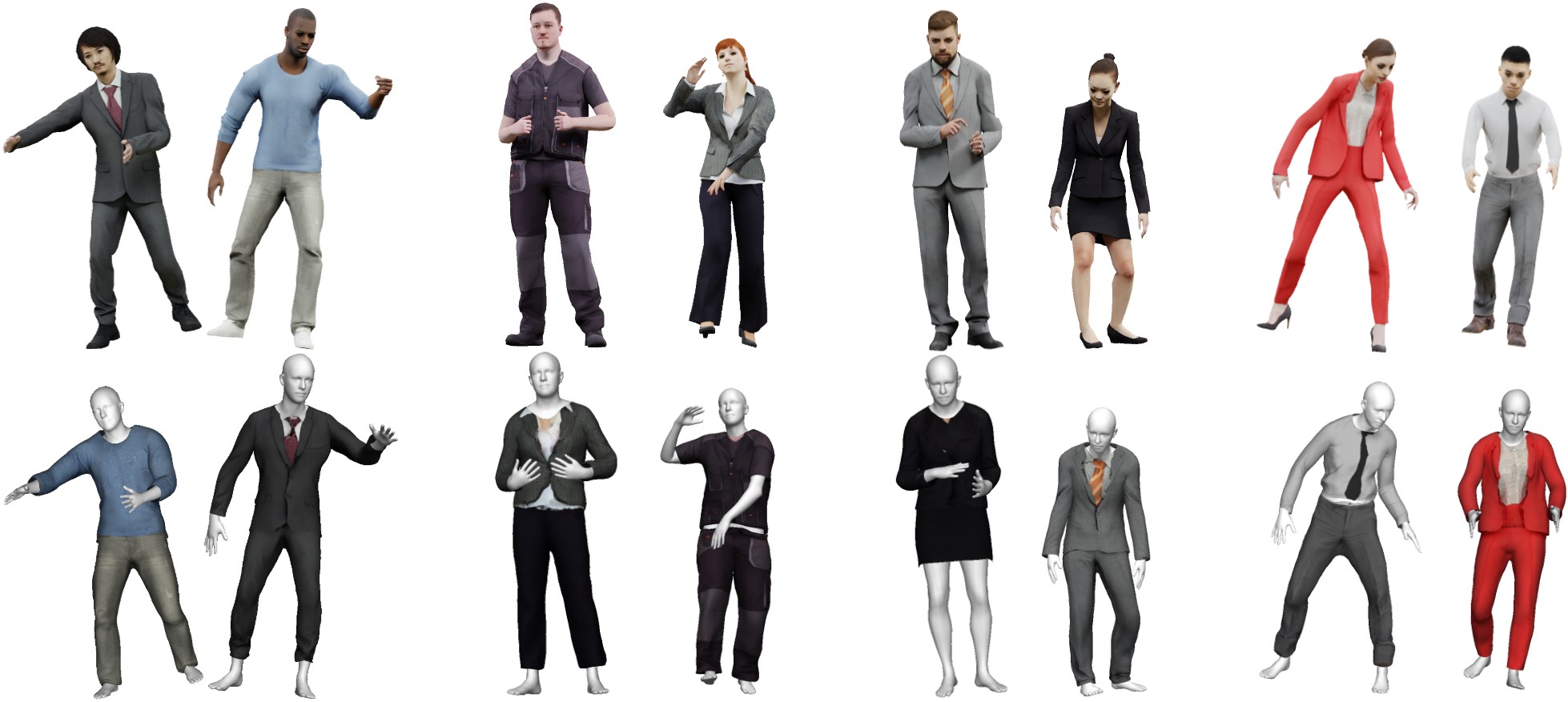}
\end{center}
    \caption{Examples of garment switching. We supply pairs clothed body images in the first row and present predicted bodies and switched garments with texture in the second row.}
\label{fig:gar_switching}
\end{figure}

In this section, we show more evaluations and results of our approach and comparisons with the state-of-the-art methods. First, we show samples of our constructed dataset in Fig.~\ref{fig:gt_data}. Then, we show more evaluations of our skinning weight network. Next, we show a qualitative comparison with MGN~[7] in Fig.~\ref{fig:compare} and more reconstruction results in Fig.~\ref{fig:HD_test}, Fig.~\ref{fig:DW}, and Fig.~\ref{fig:real}. Finally, more garment transfer and switching results are presented in Fig.~\ref{fig:exchange}, Fig.~\ref{fig:gar_transfer} and Fig.~\ref{fig:gar_switching}.

\textbf{Samples of Constructed Dataset.} In Fig.~\ref{fig:gt_data}, we present more samples of our constructed Synthetic Dataset and HD Texture Dataset. Our constructed datasets include various kinds of postures and garments, and the garment geometries match quite well with the corresponding images.

\begin{table}
\begin{center}
\caption{The skinning weight $\ell_{1}$ errors($\times10^{-3}$) between the predicted weight by our skinning weight network and ground truth, and the Euclidean distance(ED) in mm between deformed meshes with predicted and ground truth weights.}
\label{tab:rec_ws}
\begin{tabular}{|c|c|c|c|c|}
\hline
type &$\ell_{1}$ mean & $\ell_{1}$ std & ED mean & ED std  \\
\hline
l-shirt & 0.81 & 2.83 & 0.40 & 0.57 \\
s-shirt & 0.84  & 2.82 & 0.43 & 0.54  \\
l-pant & 0.45 & 1.92 & 0.30 & 0.36 \\
s-pant & 0.51 & 2.23  & 0.34 & 0.61 \\
l-skirt & 0.71 & 2.88 & 0.64& 2.00 \\
s-skirt & 0.66 & 2.55 & 0.42& 0.90 \\
\hline
\end{tabular}
\end{center}
\end{table}

\begin{figure}
\begin{center}
\includegraphics[width=\linewidth]{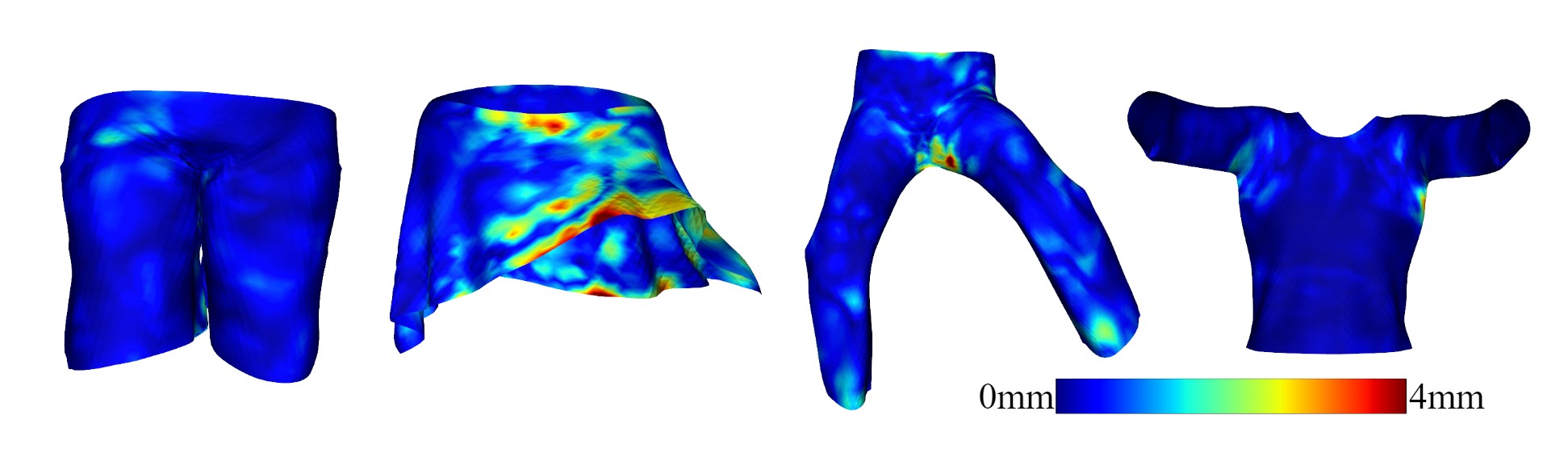}
\end{center}
   \caption{Error maps between deformed clothes with predicted skinning weights and ground truth weights.}
\label{fig:ws_errormap}
\end{figure}

\textbf{Evaluation of Skinning Weight Network.} We test the reconstruction ability of our skinning weight network. For each neutral garment of the whole test set, we reconstruct the skinning weights with our network and compute the $\ell_{1}$ error relative to the ground truth weights. Then, to evaluate skinning deformation, we compute the Euclidean distances between deformed garments with predicted and ground truth weights, respectively. Twenty random postures from the Mocap dataset are used to deform these neutral garments. From second to fifth columns of Table~\ref{tab:rec_ws}, the average and standard deviation of $\ell_{1}$ error and Euclidean distance for each garment type are given in turn. As can be observed from the results, our model can generate highly accurate results. In Fig.~\ref{fig:ws_errormap}, we visualize the error maps of several deformed garments of different types. 

\begin{figure}
\begin{center}
\includegraphics[width=\linewidth]{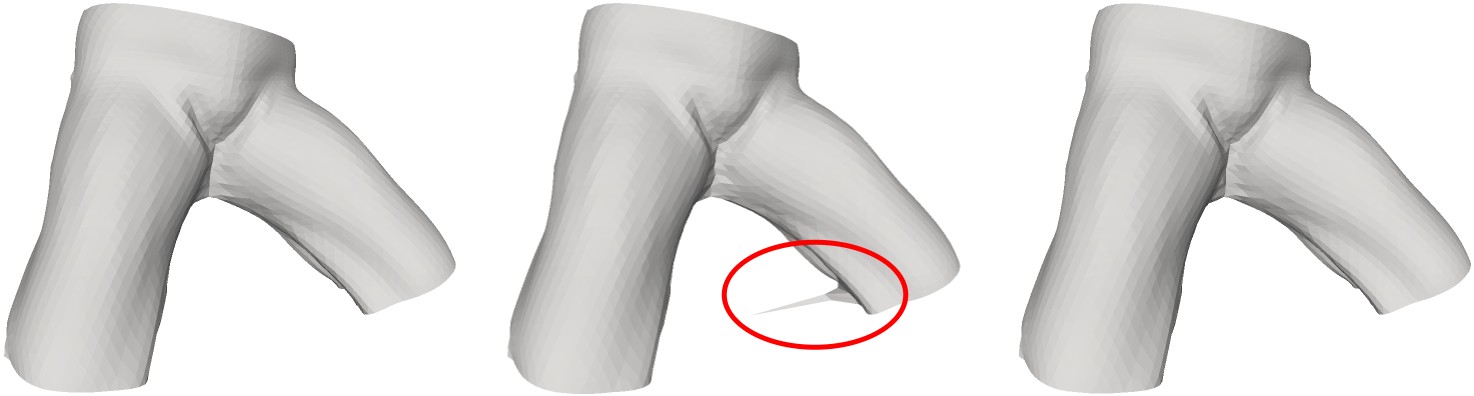}
\end{center}
   \caption{Ablation study of the vertex normals as input for skinning weight network. On the left is the deformed pant with GT weights, the middle is the result of the network without normals as input, and the right is the result of the network with full input. Predictions without normals as input would produce artifacts in ambiguous vertices.}
\label{fig:ws_norms}
\end{figure}

For some specific garments, close vertices in some parts may have very different skinning weights. For example, the vertices on either side of the crotch have close vertex positions, while the skinning weights are very different because they belong to different legs. In this case, the vertex normals of input could supply useful information to distinguish these ambiguous vertices. As shown in Fig.~\ref{fig:ws_norms}, the network without normals as input can not predict accurate weights of some vertices, leading to significant errors and artifacts.

\textbf{Qualitative Comparison with MGN.}  We show two comparison results on BUFF~[56] dataset with MGN~[7] in Fig.~\ref{fig:compare}. The results by MGN are generated by 8 views input without postprocessing optimization. We can see that the predicted body shapes by our method match the input image more closely. Besides, our predicted garments can capture size variation, such as the length of trouser legs. For the MGN method, the predicted garments are bound with SMPL, which limits its expression ability of large displacement. Therefore, although the input images are different, the results of MGN tend to maintain the same style and size for the same garment type.

\begin{figure}
\begin{center}
\includegraphics[width=\linewidth]{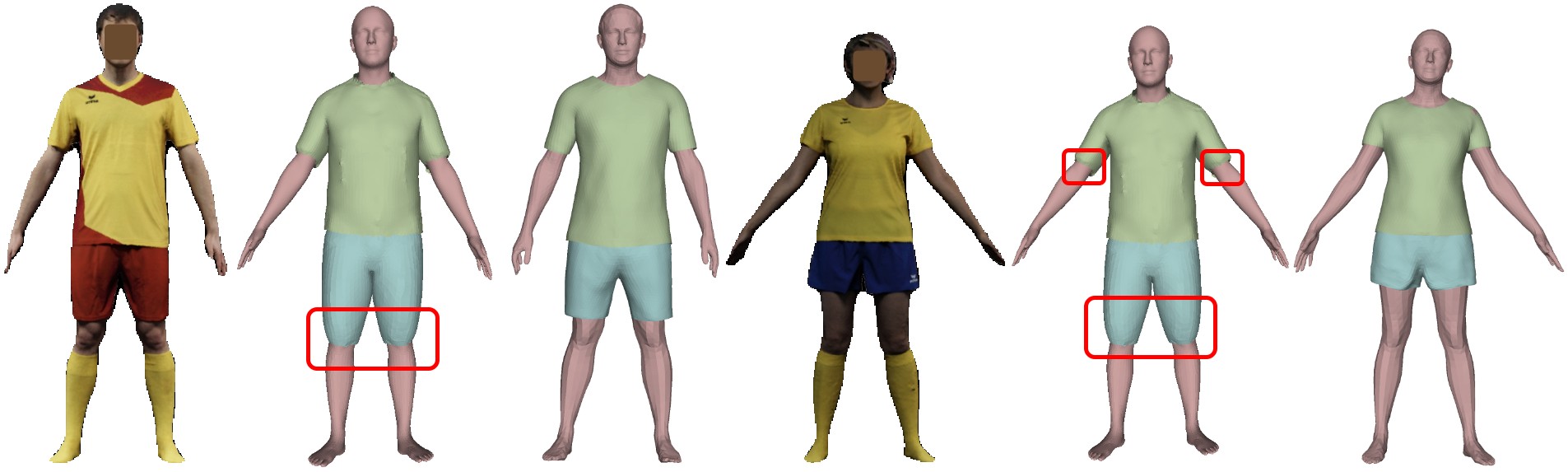}
\end{center}
   \caption{
   Two comparison results with MGN on BUFF. From left to right of each group, the reference image, predicted shapes(without post-optimization) by MGN, and predicted shapes by our method are presented. We can observe that the predicted body shapes by our method match the input images better than the ones by MGN. Moreover, our garments can capture the size variation of the input image, while predicted garments of MGN tend to maintain similar size for the same garment type even with different input images. The mismatched parts between predicted garment size and the input image are marked with red boxes.}
\label{fig:compare}
\end{figure} 

\textbf{Qualitative Results.} To demonstrate the reconstruction ability of our method, we show more results on different datasets. In Fig.~\ref{fig:HD_test}, we show some reconstruction results from our  test set. In Fig.~\ref{fig:DW}, we show some reconstruction results from the Digital Wardrobe~[7]. In Fig.~\ref{fig:real}, reconstruction results on real images are given. From these results, we can see that our method can capture the body and garment shapes from the input images quite well.

\textbf{Garment Transfer.} Fig.~\ref{fig:exchange} shows a garment switching example and two garment transfer examples between images. All the reconstructed shapes are rendered with texture to improve the visual authenticity. In Fig.~\ref{fig:gar_transfer} and Fig.~\ref{fig:gar_switching}, we present more results of garment transfer and switching on HD Texture test dataset. Based on our method, we can easily transfer or switch garments geometry and texture between two images, even with different garment types. These applications further confirm the ability of our method to correctly predict the 3D shapes of garment and human body.

\section{Limitations}

\begin{figure}[t!]
\begin{center}
\includegraphics[width=\linewidth]{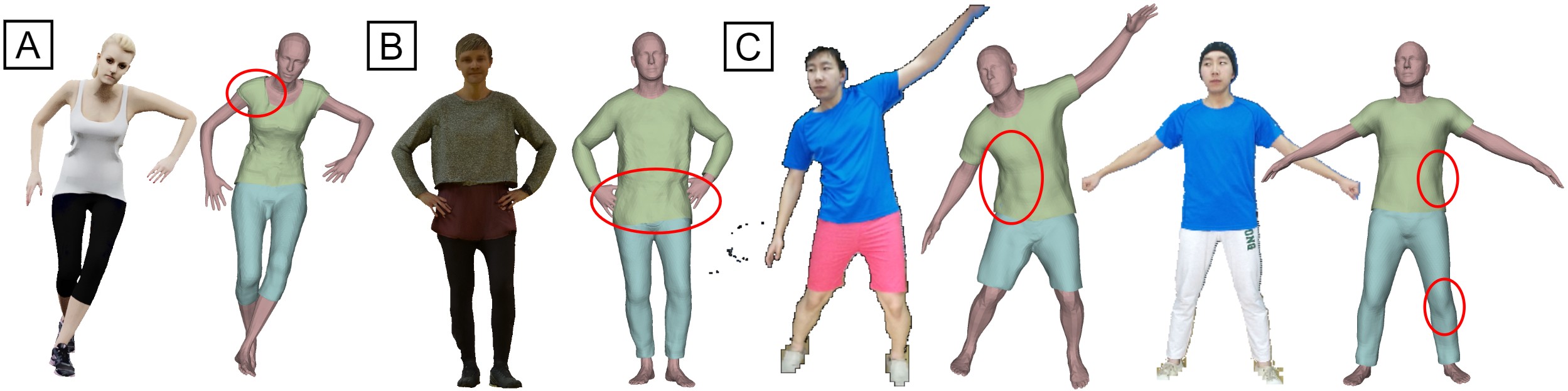}
\end{center}
    \caption{The proposed method has shortcomings. We present three challenges for future work in the figure. A) Incorrect garment reconstruction due to unsupported garment type. B) Multi-layered garments are treated as a single layer garment. C) Over-smooth results for two images captured by Kinect v2 camera~[55].}
\label{fig:limitation}
\end{figure}

Our method still has some limitations which deserve further study.
\begin{itemize}
\item Our method currently supports six garment types. Therefore, current trained model can not correctly predict the garment type which does not belong to the six garment types. One example is shown in Fig.~\ref{fig:limitation} A. However, our method can be extended to support new garment type with the same strategy in the paper.

\item In this work, our method can recover the body and garment shapes while we does not consider the hair, shoes, hats, and multi-layered clothing. We show an example of multi-layered clothing in Fig.~\ref{fig:limitation} B.
\item Clothed body images contain enormous diversities in the aspect of cloth types, textures, body shapes, lighting conditions, background, and camera angles. The trained model might produce over-smooth results for test images that have very different styles with the training dataset. Two examples are shown in Fig.~\ref{fig:limitation} C. However, synthesizing more realistic data and utilizing more real data in our proposed framework can alleviate this problem.

\end{itemize}

\end{document}